\documentclass{article}

    \usepackage[nonatbib,preprint]{neurips_2023}

\usepackage[utf8]{inputenc} 
\usepackage[T1]{fontenc}    
\usepackage{epigraph}
\usepackage{algorithmic}
\usepackage{enumitem}
\usepackage{graphicx}
\usepackage{algorithm}
\usepackage{wrapfig}
\usepackage{xcolor}
\usepackage{url}            
\usepackage{booktabs}       
\usepackage{amsfonts}       
\usepackage{amsmath}
\usepackage[page,header]{appendix}
\usepackage{titletoc}
\usepackage{amsthm}
\usepackage{lscape} 
\usepackage{amssymb}
\usepackage{multirow}
\usepackage{nicefrac}       
\usepackage{microtype}      
\usepackage{titlesec}
\usepackage{makeidx}
\usepackage[skip=8pt]{caption}
\usepackage[hidelinks]{hyperref}

\newcommand{\specialcell}[2][c]{%
\begin{tabular}[#1]{@{}c@{}}#2\end{tabular}}

\titleformat{\subsection}[runin]{\normalfont\bfseries}{\thesubsection}{0.5em}{}

\titleformat{\subsubsection}[runin]{\normalfont\bfseries}{\thesubsubsection}{0.5em}{} 

\titlespacing{\section}{0em}
  {0.1em}
  {0.1em}

\titlespacing{\subsection}{0em}
  {0.1em}
  {0.2em}

\titlespacing{\subsubsection}{0em}
  {0.1em}
  {0.2em}

\title{Im-Promptu: In-Context Composition \\ from Image Prompts}

\author{%
  Bhishma Dedhia\textsuperscript{1}, Michael Chang\textsuperscript{2}, Jake C. Snell\textsuperscript{3}, Thomas L. Griffiths\textsuperscript{3,4},  Niraj K. Jha\textsuperscript{1}\\
  \\
\textsuperscript{1}Department of Electrical and Computer Engineering, Princeton University\\
\textsuperscript{2}Department of Computer Science, University of California Berkeley\\
\textsuperscript{3}Department of Computer Science, Princeton University \\
\textsuperscript{4}Department of Psychology, Princeton University\\
  \texttt{\{bdedhia,js2523,tomg,jha\}@princeton.edu},  \texttt{mbchang@berkeley.edu}\\
}

\begin{document}

\maketitle

\begin{abstract}

Large language models are few-shot learners that can solve diverse tasks from a handful of demonstrations. This implicit understanding of tasks suggests that the attention mechanisms over word tokens may play a role in analogical reasoning. In this work, we investigate whether analogical reasoning can enable in-context composition over composable elements of visual stimuli. First, we introduce a suite of three benchmarks to test the generalization properties of a visual in-context learner. We formalize the notion of an analogy-based in-context learner and use it to design a meta-learning framework called \textit{Im-Promptu}. Whereas the requisite token granularity for language is well established, the appropriate compositional granularity for enabling in-context generalization in visual stimuli is usually unspecified. To this end, we use \textit{Im-Promptu} to train multiple agents with different levels of compositionality, including vector representations, patch representations, and object slots. Our experiments reveal tradeoffs between extrapolation abilities and the degree of compositionality, with non-compositional representations extending learned composition rules to unseen domains but performing poorly on combinatorial tasks. Patch-based representations require patches to contain entire objects 
for robust extrapolation. At the same time, object-centric tokenizers coupled with a cross-attention module generate consistent and high-fidelity solutions, with these inductive biases being particularly crucial for compositional generalization. Lastly, we demonstrate a use case of \textit{Im-Promptu} as an intuitive programming interface for image generation.

\end{abstract}

\section{Introduction}

\label{sec:intro}

\epigraph{No thought can be formed that isn’t informed by the past; or, more precisely, we think only thanks to 
analogies that link our present to our past.}{D. Hofstadter and E. Sander, Surfaces and Essences \cite{hofstadter2013surfaces}} 

Humans represent complex concepts by combining and organizing simpler concepts \cite{Lake2011OneSL,lakecompositonal}. 
This endows us with an uncanny skill of constructing an unlimited number of concepts by composing a relatively small 
set of previously learned basic building blocks, an ability more formally called \textit{compositional generalization}. 
For example, we can generate sentences by combining a dictionary of words in different ways and solve a wide range of 
mathematical problems using a small set of basic arithmetic operations and variables. In the case of the visual world, 
objects form naturally composable entities and object-centric primitives can be flexibly combined using abstract 
syntactic rules to yield novel and counterfactual scenes. 


The growing body of work in object-centric learning methods \cite{monet_burgess,greff_iodine,locatello_slot_att} enable the extraction of latent object representations from perceptual scenes. However, slot composition methods have been limited to \textit{ad-hoc} composition rules based on hard-coding slots into concept libraries \cite{singh_SLATE, jiang2023objectcentric}. 
A central challenge in learning a generative 
model that explicitly models the compositional grammar of the visual world is that the seemingly unlimited number of 
composition rules that undergird real-life visual entities prevents the \textit{explicit} specification of each rule. For example, a warm spring day consists of blooming flowers and freshly verdant expanses, but come autumn, the same scene stipulates the presence of golden meadows carpeted with fallen leaves.  State-of-the-art text-to-image models 
\cite{dalle, dalle2,imagen,stable_diff} circumvent this bottleneck by grounding images in natural language. This 
aligns the representations of entities in an image to the compositional structure of language. While the visual quality 
of these models is stunning, they often require engineering complex language prompts to elicit the desired output. 
They also suffer from limited understanding of object relations \cite{marcus_dalle}. Moreover, studies 
\cite{xu_numbersense} have shown that infants as young as six months old can extract compositional structure from 
sequences of visual events, even before they acquire language.  Such compositional abstraction independent of language faculty exhibited by humans remains elusive for contemporary models.


\begin{wrapfigure}{l}{0.5\linewidth}
  \centering
  \includegraphics[width=\linewidth]{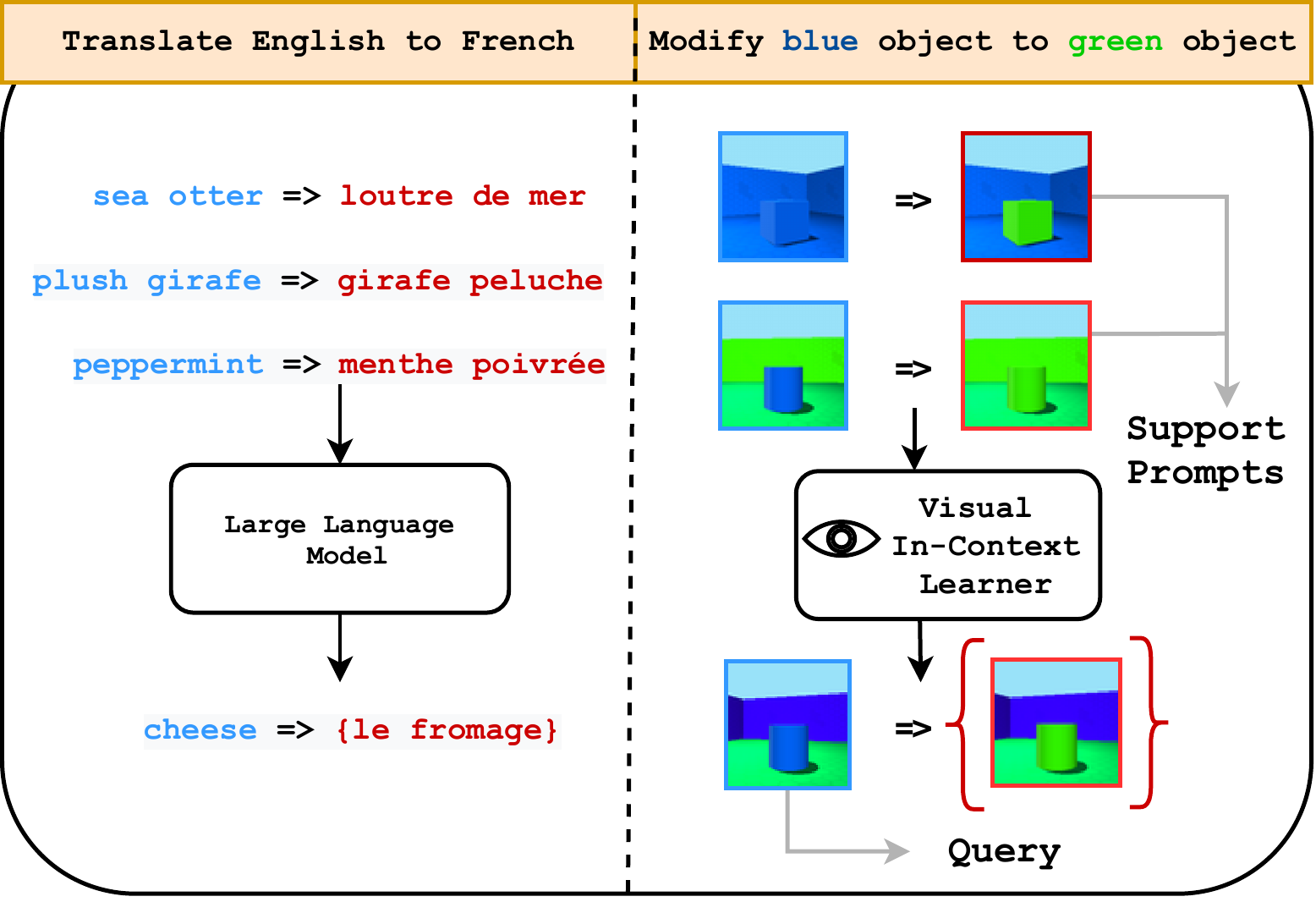}
  \caption{LLMs perform in-context understanding of a task from a few examples in an analogical manner (left).  We use 
analogy-making to implicitly understand the composition rules over visual stimuli made of object-like entities (right).
  }
  \label{fig:hook}
\end{wrapfigure}

Large language models (LLMs) \cite{brown2020language, gpt-j, chowdhery_PALM, jurrasic, bloom, openai2023gpt4}, on the 
other hand, demonstrate \textit{implicit} composition skills. Pre-trained on large text corpora, these models have 
impressive emergent few-shot learning capabilities. They can solve novel tasks on the fly without gradient updates 
by following examples shown in the instruction prompts, a process known as \textit{in-context learning}. Under the 
hood, in-context learning uses the inherently compositional nature of the input text to implicitly infer the task structure and how to apply the task structure to a query 
\cite{wei2023larger}. In the most simple form, the prompt-based few-shot learning paradigm can be formalized as 
\textit{analogy solving} of the type $A:B::C:D$, where \textit{A} represents an example task, \textit{B} represents 
the example solution, and \textit{C} denotes the query. The generated solution is the analogy completion $D$. Analogies over compositional elements, therefore, provide a direct isomorphism between in-context learning for language and 
learning implicit composition rules for visual objects, as depicted in Fig.~\ref{fig:hook}. While LLMs use pre-specified tokenizers \cite{sennrich2016neural,wordpiece,kudo2018sentencepiece}, the compositional granularity for \textit{implicit} visual understanding is unspecified. On the other hand, slot methods induce object-level representations but fail to learn composition rules. In this work, we then ask:

\textit{Can analogy-solving enable in-context generalization over object-centric compositional elements of visual entities? }

While many linguistic corpora have been extracted from the Internet for text-based tasks, to the best of our knowledge, 
no datasets exist for measuring in-context composition capability from visual prompts. To this end, we introduce a 
suite of three diverse benchmarks in Section \ref{sec:benchmark} built on top of rich combinatorial spaces to 
systematically test image-based in-context learning abilities: (a) 3D Shapes \cite{3dshapes18}, (b) BitMoji Faces, 
and (c) CLEVr Objects \cite{clevr_johnson}. If the attention over word tokens enables analogical understanding in LLMs, 
can we transfer this mechanism to visual analogies?  We formulate a unifying framework in Section \ref{sec:framework} 
to formalize this notion. We present a visual analogy-based meta-learning algorithm called \textit{Im-Promptu} in 
Section \ref{sec:impromptu} and use it to train agents with a cross-attention module. How does one tokenize visual 
inputs? To answer this, we model several generative agents in Section \ref{sec:agents}, encompassing a continuum 
ranging from strong compositional learners using object-centric abstractions to non-compositional monolithic 
representation learners. Extensive experiments presented in Section \ref{sec:experiments} show the dependence of 
robust in-context learning for visual stimuli on object-centric slots as visual tokenizers and cross-attention. 
We demonstrate the use of our visual analogy framework in task extrapolation (Section \ref{sec:source_target}), 
combinatorial generalization (Section \ref{sec:composite}), and generating counterfactual images 
(Section \ref{sec:prompt}).

\section{Related Work}
In this section, we discuss prior works and draw connections to relevant research areas.

\subsection{Object-Centric Learning:} This growing body of work targets unsupervised decomposition of scenes into a 
set of object-centric vectors from raw perception data. A common approach involves modeling an autoencoder consisting 
of a latent factorization of object slots that are independently decoded to reconstruct the scene 
\cite{monet_burgess, greff_iodine, locatello_slot_att,greff_slot,engelcke_genesis,engelcke_genesisv2,kipf_savi,saviplus_elsayed,singh_STEVE}. These methods 
effectively segment scenes into representations of underlying objects but do not learn the flexible combination of 
object representations. On the other hand, the authors in \cite{singh_SLATE} replace the independent slot decoding 
bias with an Image-GPT \cite{chen_imageGPT} and demonstrate novel slot composition. While this work largely answers 
how to compose a \textit{given} set of visual entities, their slot selection is performed manually through organization 
of slots into clustered libraries. In our work, we use a meta-learning approach to help the learner implicitly 
understand the composition task. Recently, the authors of \cite{chang_implicit} recast the iterative slot refinement 
problem as finding of fixed points through implicit differentiation to improve the stability and computational cost of 
the slot-attention (SA) technique. The authors of \cite{constellation} use specific inductive biases to disentangle 
object contents from spatial information that enables relational reasoning. Our work produces a more generalized 
composition beyond spatial relations. 

\subsection{In-Context Learning in LLMs:}

LLMs \cite{brown2020language,  gpt-j, chowdhery_PALM, jurrasic, bloom, openai2023gpt4} can perform in-context learning at inference time  from natural language prompts. In this work, we aim to extend this paradigm beyond language without language grounding.  Empirical studies \cite{min_incontext, chan_incontext} have found that the success of in-context learning critically hinges on idiosyncrasies in training distribution, prompt text structure, and label examples. In contrast, theoretical interpretations \cite{xie_bayesian} have formulated it as an instance of implicit Bayesian inference. Prior works \cite{chen_meta,min_meta} have also attempted to improve language models by meta-learning on an in-context learning objective. 

\subsection{Analogical Reasoning:}

The ability to make analogies has been shown to have important implications in various domains, such as problem-solving, 
creativity, learning, and counterfactual thinking 
\cite{guilford1950creativity,GENTNER1983155,markman1993structural,holyoak1995mental,hofstadter2013surfaces}.
Various models have been proposed to explain the underlying mechanisms of analogy-making. Popular symbolic approaches 
include ARGUS \cite{reitman1965cognition}, Transfer Frames \cite{WINSTON1978147}, and Structural Mapping Engine 
\cite{Falkenhainer1989}, connectionist models include LISA \cite{Hummel1997} and Star-1 \cite{Halford1994}, and models 
that occupy a middle ground like Copycat \cite{Hofstadter1995}. Contemporary deep-learning models 
\cite{bongard,rpm,hill_LABC} attempt to solve Raven's Progressive Matrices and Bongard Problems, types of visual 
analogies present in IQ tests. The authors of \cite{reed_dvm} propose a generative deep visual analogy-making model 
based on a parallelogram regularizer on the latent space to disentangle variation factors. Besides, several works \cite{mikolov_word2vec,webb_emergent} show emergent analogical structure in language models.

\subsection{Compositional Generative Models:}

Prior works \cite{du_eb, liu_diffusion} use energy-based models to represent composable factors of variation of a 
scene. However, these methods use concept labels \cite{hudson2021compositional} or are grounded in text and are limited to a few concept composition 
operations. A follow-up work \cite{du_unsupervised_ebm} describes unsupervised energy-based concepts; however, the 
re-composition is performed via manually-picked concepts. Recent works in text-to-image models use language grounding 
to induce compositionality. DALL-E \cite{dalle} uses language modeling over the joint space of the text and image 
tokens to facilitate image generation. The latest text-to-image models \cite{dalle2,imagen,stable_diff} use diffusion 
modeling to guide the image generated from the text. A growing body of work \cite{prompt_editing,unitune,imagic} 
explores text-grounded controllable image editing via prompt engineering, where most parts of the image are preserved, 
but specific objects are re-composed.

\subsection{Modular Representations and Attention:} 
A swath of recent works proposes neural methods that combine modular primitives with recurrent mechanisms and cross-attention. The authors of \cite{goyal2020recurrent} propose the Recurrent Independent Mechanisms (RIMs) to learn weakly coupled dynamical systems with independent compositional neural modules that communicate via a sparse attention bottleneck. Another work \cite{goyal2020object} uses object-centric abstractions as `files' to store factorized declarative knowledge and the dynamics of the environment.  A follow-up work \cite{goyal2022neural} introduces neural abstractions of the classical Production Systems framework to learn rule-based templates, thereby facilitating compositional dynamics of the environment that can quickly adapt to novel stimuli. Most recently, the authors in \cite{ke2022learning} propose an attention-based architecture to learn in-context causal relationships between entities.



\section{Benchmarks}
\label{sec:benchmark}
\begin{figure}[!htb]
    \centering
    \includegraphics[width=\linewidth]{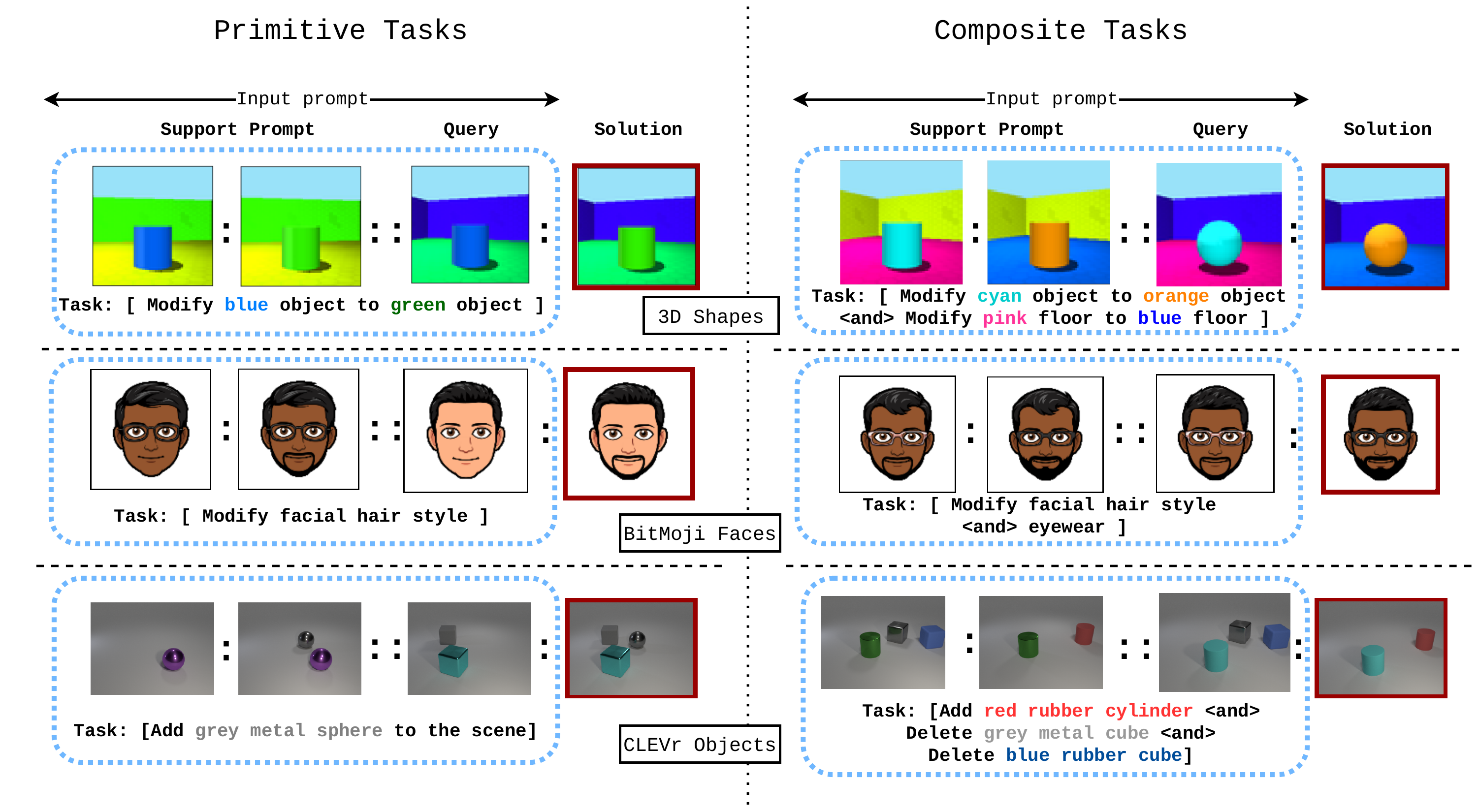}
    \caption{We use combinatorial image spaces to set up visual in-context learning tasks. Each task requires the generation of a solution from a query and supporting examples. The latent description of the task is described within the square 
parentheses. Left: Primitive tasks modify a composable element in isolation. Right: Composite tasks combine primitives to test the combinatorial generalization skills of the learner.}
    \label{fig:benchmarks}
\end{figure}
Humans can contextualize past visual stimuli in the face of everyday experiences and even use them for imagining 
counterfactuals. Having seen an oak table, we can, with little difficulty, look at a metal armchair and understand the 
concept of an oak armchair. How do we then systematically test such abilities of a visual in-context learner? In 
this work, we posit that an answer to this question lies in combinatorial visual spaces with controllable factors of 
variation, which provide a \textit{semantically rich, simple, scalable, and composable} avenue for setting up such 
tasks. We create a suite of three benchmarks (Fig.~\ref{fig:benchmarks}) from compositional image creators that 
include (a) 3D Shapes \cite{3dshapes18}, (b) BitMoji Faces, and (c) CLEVr Objects \cite{clevr_johnson}. Each benchmark 
has a split of primitive training tasks and out-of-distribution test tasks. The \textit{primitive} tasks 
(Fig.~\ref{fig:benchmarks}, left) constitute an analogy where a \textit{single} composable element of the visual 
stimulus is modified in isolation.  The description of each benchmark with its primitive set of tasks is given next:

\subsection{3D Shapes Image Prompts:} This dataset consists of static scenes of various objects lying on a colored 
floor in front of a colored wall viewed from different orientations. To set up primitive tasks, we consider the four 
properties $P$ = \{\texttt{object} \texttt{color}, \texttt{wall} \texttt{color}, \texttt{floor}  \texttt{color}, 
\texttt{scene} \texttt{orientation}\}, where each property can take multiple values within a domain. For each such 
task, the latent relation $r(.)$ for the analogy prompt takes a property $p \in P$ and modifies it from a 
\textit{source} domain value to a \textit{target} value. This benchmark comprises $80000$ such tasks with a maximum 
of four supporting image pairs for each instance and with a roughly equal split across properties.
\subsection{BitMoji Faces Image Prompts:} BitMoji is an avatar creator service for social media users that allows them 
to create intricate cartoon faces. We queried the BitMoji API \cite{bitmoji} to collect faces with four underlying 
dynamic elements $P =$ \{\texttt{skin} \texttt{tone}, \texttt{facial} \texttt{hair} \texttt{style}, \texttt{hair} 
\texttt{style}, \texttt{eyewear}\}. Much like the previous case, primitive tasks are generated by intervening on a 
\textit{source} value of a property and modifying it to a \textit{target} value. We populate the training set with 
80000 tasks, with four demonstrations available per task.

\subsection{CLEVr Objects Image Prompts:} CLEVr is a popular visual question-answering dataset with the visual component 
consisting of multiple objects lying in a scene. We use the CLEVr rendering engine to set up primitive tasks that 
include \texttt{adding} and \texttt{deleting} the same object across various scenes. In this fashion, we 
generate $60000$ tasks with three examples per task.

A composable image space allows the concurrent modification of individual visual elements. One can construct 
\emph{composite} 
tasks (Fig.~\ref{fig:benchmarks}, right) out of the combination of primitives shown in the training set. A $k$-composite 
task 
is denoted as $R^k(.)=r_1 \circ r_2 \circ \cdots \circ r_k$ where $r_1, \cdots , r_k \in \mathcal{R}$. We provide details of the $k$-compositeness test of each of our agents in Section \ref{sec:composite}. Finer details about each benchmark, along with visual examples from every task, are detailed in Appendix \ref{app:benchmarks}. We will release the complete dataset of primitive and composite tasks upon the publication of this article.

\section{Methods} 
 To learn compositional structure over visual stimuli, we begin by formulating a unified mechanistic interpretation of an in-context compositional learner using insights from design elements of LLMs.
\subsection{In-Context Learning as Analogy Completion:}
\label{sec:framework}
A general \textit{compositional} in-context learner $\mathcal{M_{\phi,\alpha,\theta}(.)}$ can be formalized via
the following three key components: 
\begin{itemize}[nolistsep,noitemsep]
    \item  An encoder $\mathcal{E}_\phi(.)$ that maps the input space $ \mathcal{X} $ to the compositional space $\mathcal{C}$ $\triangleq \mathcal{E}_\phi(.): \; \mathcal{X} \longmapsto \mathcal{C}$
    \item An executor $\mathcal{T_\alpha(.\;,.\;,.)}$ that maps compositional entities from example tasks, example solutions, and a query task to the query solution $\triangleq \mathcal{T}_\alpha(.,.,.): \; \mathcal{C} \times \mathcal{C} \times \mathcal{C}  \longmapsto \mathcal{C}$
    \item A decoder  $\mathcal{D}_\theta(.)$ that maps the compositional space back to the input space  $\triangleq \mathcal{D}_\theta(.): \; \mathcal{C} \longmapsto \mathcal{X}$ 
\end{itemize}
Coupling the above functions together yields the learner:
\begin{gather}
    \mathcal{M} \triangleq \mathcal{D_\theta}\left(\mathcal{T}_\alpha\left(\mathcal{E}_\phi(.),\mathcal{E}_\phi(.),\mathcal{E}_\phi(.)\right)\right): \mathcal{X} \times  \mathcal{X} \times \mathcal{X}\longmapsto \mathcal{X}
\end{gather}
The inclusion of $\mathcal{E}_\phi(.)$ and $\mathcal{D}_\theta(.)$ is crucial for compositional abstraction since
the input space may not be \textit{a priori} composable as in the case of images. Beyond this, we do not place any
strong parametric constraints on the model. Let $\mathcal{R}$ denote a task set. For any task $r \in \mathcal{R}$, 
define the task analogy $\mathcal{\zeta}$ over input pairs $x_1,x_2$ as $x_1:r(x_1)::x_2:r(x_2) \triangleq A:B::C:D$. An effective in-context learner over $\mathcal{R}$ then satisfies the following property \cite{garg2022}:
\begin{gather}
\label{eq:in_context}
    \forall r \in \mathcal{R},\:\mathbb{E}_{\mathcal{\zeta} \sim \mathcal{P}(r,\mathcal{X})}\left[\mathcal{L}\left(D;\mathcal{M}_{\phi,\alpha,\theta}(A,B,C)\right)\right] \leq \epsilon 
\end{gather}
Here, $\mathcal{P}$ denotes the distribution of the analogies resulting from the input space $\mathcal{X}$ and 
task $r(.)$, $\mathcal{L}(.)$ denotes a loss metric, and $\epsilon$ is an error bound. The above property simply
states that a good in-context learner solves analogies defined over a task set within some finite error bound. In
Appendix \ref{app:LLMs}, we describe LLMs as an instantiation of this framework.

\subsection{Im-Promptu Learning:}
\label{sec:impromptu}

Pre-training LLMs on large-scale text data leads to the emergence of highly transferable internal representations that allow them to adjust to novel tasks by simple priming
on a few demonstrations. However, it is still unclear whether the underlying structure of the training data, 
architecture, or inductive biases instilled by choice of pre-training learning algorithms cause these intriguing
properties to emerge. How can we encourage the emergence of such a general-purpose representation for visual data?
In this work, we take an explicit approach based on the formalism presented in the previous section. Having established the desideratum (Eq. \ref{eq:in_context}), we set up a learning procedure called \textit{Im-Promptu} that trains the model to generate analogy completions from image prompts. To this end, given a task set $\mathcal{R}$, in each minibatch, we sample an analogy $A:B::C:D$ that follows a latent primitive $r(.) \in \mathcal{R}$. Given a loss criterion, $\mathcal{L}$,  we make the following stochastic gradient updates of the model parameters:
\begin{gather}
    \{\phi^{t+1} , \alpha^{t+1}, \theta^{t+1}\} = \{\phi^{t} , \alpha^{t}, \theta^{t}\}\; -  \frac{\alpha}{N} \sum_{i=1}^N \nabla_{\theta^t, \alpha^t, \phi^t} \mathcal{L}\left(D,\mathcal{M}_{\phi^t,\alpha^t,\theta^t}\left(A,B,C\right)\right)
\end{gather}
 where the step size $\alpha$ is a hyperparameter. The full \textit{Im-Promptu} algorithm is laid out in Algorithm \ref{alg:training}.
 
\label{app:training}
\begin{algorithm}[!hbt]
\caption{Im-Promptu Learning Algorithm}
\label{alg:training}
\begin{algorithmic}
\STATE {\bfseries Require:} Task set $\mathcal{R}$,  Step Size $\alpha$
\STATE Initialize parameters $\theta,\alpha,\phi$ of model $\mathcal{M}$
\WHILE{not done}
\STATE Sample primitive task $r(.) \sim \mathcal{R}$
\STATE Sample $N$ input image pairs $X^{1:N}_1, X^{1:N}_2 \sim \mathcal{P}_{train}$
\FOR{$i$ in $1 \cdots N$}
\STATE Sample analogy $\zeta^i \triangleq x^i_1:r(x^i_1)::x^i_2:r(x^i_2) \triangleq A:B::C:D$
\ENDFOR
\STATE Update  $\{\theta,\alpha,\phi \}= \{\theta, \alpha, \phi\} - \frac{\alpha}{N} \nabla_{\theta, \alpha, \phi} \sum_i \mathcal{L}_\mathcal{M}(\zeta^i)$ 
\ENDWHILE
\end{algorithmic}
\end{algorithm}

\section{Background}
We give brief details of the Slot Attention Mechanism and Slot Attention Transformer (SLATE) that form essential
components of our best-performing agents.
 
\subsection{Slot Attention:}
Object-Centric Learning frameworks decompose scenes into compositional object files called slots. SA 
\cite{locatello_slot_att} is a powerful autoencoder-based inductive bias for learning such slots. The encoder 
$f_\phi(.)$ initializes a set of $N$ symmetric and independent slots $S^0_{1:N} \triangleq \{s_1^0,s_2^0,\cdots,s_N^0\}$  
that iteratively compete over $T$ timesteps to attend to the input scene $X$ and break symmetry to yield the final 
slots $S^T_{1:N}\triangleq \{s_1^T,s_2^T,\cdots,s_n^T\}$. A decoder $g_\theta \left(.\right)$ then individually 
decodes each slot in the final set into the pixel space to yield $N$ images $I_{1:N}$ and their corresponding masks 
$M_{1:N}$. A convex combination of the images, using masks as weights, reconstructs the scene $\hat{X}$. 
\begin{gather}
S^T_{1:N} = f_\phi(S^0_{1:N},X) \\
I_1, M_1 = g_\theta(s_1^T), \cdots, I_N, M_N = g_\theta(s_N^T)\\
 \hat{X} = M_1 \times I_1 + \cdots + M_N \times I_N
\end{gather}

\subsection{Slot Attention Transformer:}
\label{sec:slate}
Recent work \cite{singh_SLATE} solves the pixel independence problem in SA \cite{locatello_slot_att} by learning
a latent discrete library from pixels using a discrete variational autoencoder (dVAE) \cite{discrete_VAE,vq_vae} 
and then running SA over the vocabulary space of the dictionary. An imageGPT decoder \cite{chen_imageGPT} learns to 
predict the sequence of image latent from the slot prompts and, as a result, enables novel composition of slots. We 
lay out the forward pass of the autoencoder architecture below but defer the finer details to the original article.
\begin{gather}
Z_{1:L} = \{z^1, \cdots , z^L\} = f^{dVAE}_\phi(X) \rightarrow\text{ dVAE encoder} \\
    S^T_{1:N} = SA_\phi(Z_{1:L},S^0_{1:N}) \rightarrow \text{Slot Attention} \\ 
\hat{Z}_{1:L} = g_\theta^{GPT}(S^T_{1:N}) \rightarrow \text{ImageGPT} \\
\hat{X} = g_\theta^{dVAE}(\hat{Z}_{1:L}) \rightarrow \text{dVAE decoder}
\end{gather}
Here, $X$ is the input, $Z_{1:L}$ is the latent discrete sequence of length $L$, and $S_{1:N}$ denotes the $N$ slots. 

\section{Learning Agents} 

\label{sec:agents}
In this section, we explore various modeling choices for a visual in-context learner ranging from simple pixel-space 
rules to object-centric learners (OCLs). 

\subsection{Pixel Baseline:} This non-parametric baseline manipulates the pixel space using a simple addition rule. We denote the support prompt as $A:B$ and the query as $C$, the solution $\hat{D}$. Then the Pixel baseline predicts $\hat{D}$ as the  outcome of a linear transformation, $\hat{D} = C + (B-A)$.

\subsection{Monolithic Learner:} Our first learning agent, inspired from \cite{reed_dvm}, is entirely non-compositional and uses latent monolithic vectors to execute the image prompts as follows:
\begin{gather}
    \mathcal{V} = f_\alpha\left(\left[e_\phi(A),e_\phi(B)\right]\right) ,\;\hat{D} = d_\theta\left(h_\beta\left([\mathcal{V},e_\phi(C)]\right)\right)
\end{gather}
The encoder $e_\phi(.)$ and decoder $d_\theta(.)$ are convolutional and deconvolutional networks, respectively. The 
inference network $f_\alpha(.)$ and executor network $h_\beta(.)$ are modeled as multi-layered perceptrons. Here,
$[\;]$ denotes the concatenation operator. This architecture is detailed in Appendix \ref{app:monolithic}. 

\begin{figure}[!htb]
    \centering
        \includegraphics[width=\linewidth]{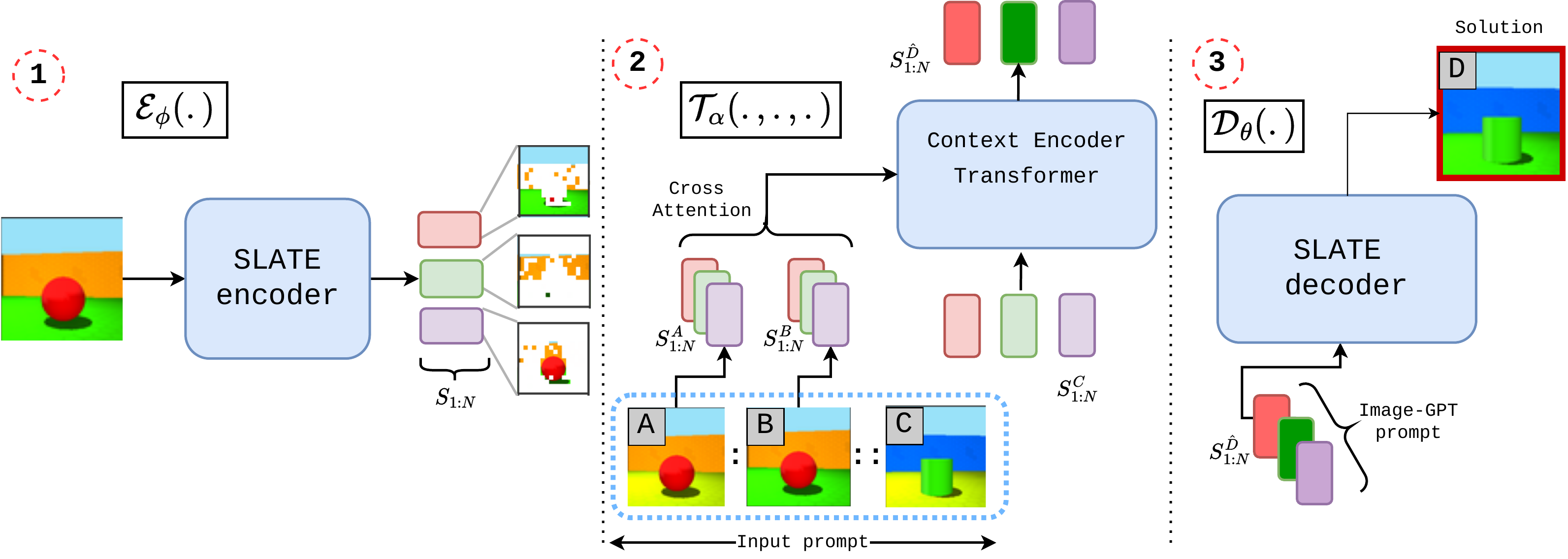}
    \caption{The OCL uses compositional object vectors to answer image prompts. $\mathbf{(1)}$
The encoder $\mathcal{E}_\phi(.)$ is a pre-trained SLATE \cite{singh_SLATE} encoder that uses a 
dVAE to obtain a latent discrete sequence for the input image and runs SA 
\cite{locatello_slot_att} over the sequence to extract object-centric slots. $\mathbf{(2)}$ The CET modifies query 
slots $S^C_{1:N}$ via cross-attention over the support prompt slots $S^A_{1:N}, S^B_{1:N}$. $\mathbf{(3)}$ The modified slots are used to prompt an Image-GPT \cite{chen_imageGPT} that outputs the latent discrete sequence of the task solution $D$ that is decoded by the dVAE decoder.} 
    \label{fig:ocl_agent}
\end{figure}

\subsection{Inpainting Model:} Inspired by recent work on inpainting for In-Context learning \cite{bar2022visual}, this agent
(Inpaint., visualized in Appendix \ref{app:inpainting}) is modeled via an autoencoder architecture that fills in missing
patches of an image. The architecture is composed of a discrete variational autoencoder (dVAE) that discretizes the input patches and a Vision Transformer \cite{dosovitskiy2021image} auto-encoder to impute missing values. While the monolithic agent separately encodes the components of the visual analogy, this model jointly represents them as a $2 \times 2$ image grid. The model is pre-trained using the masked autoencoder reconstruction task \cite{he2021masked}. Subsequently, at inference time, the analogy solution is represented as a missing image in the grid. Details have been given in Appendix \ref{app:inpainting}.

\subsection{Object-Centric Learner (OCL):} This agent, as depicted in Fig.~\ref{fig:ocl_agent}, is a strongly compositional learner that uses object-centric inductive biases. The encoder and decoder of the framework in Section \ref{sec:framework} are parametrized using SLATE \cite{singh_SLATE} (see Section \ref{sec:slate}) as follows:
\begin{gather}
    \mathcal{E}_{\phi}(.) \triangleq SA_\phi(f_\phi^{\text{dVAE}}(.)),\;\mathcal{D}_{\theta}(.) \triangleq  g_\theta^{dVAE}(g_\theta^{GPT}(.))
   \label{eq:slate_decoder}
\end{gather}
 A Context Encoder Transformer (CET) $\triangleq\mathcal{T}_\alpha$(.\;,.\;,.) (.) with interleaved layers of: $(a)$ 
cross-attention on the context slots and $(b)$ self-attention, induces sparse modifications on the query slots to complete the analogy.
\begin{gather}
    S_{1:N}^A =  \mathcal{E}_{\phi}(A),\;S_{1:N}^B =  \mathcal{E}_{\phi}(B),\;S_{1:N}^C =  \mathcal{E}_{\phi}(C) \\
    S_{1:N}^D = CET\left(\text{query =}S_{1:N}^C ,\text{keys, values =}S_{1:N}^A,S_{1:N}^B\right)
\end{gather}
The latent sequence $\hat{Z}_D$ predicted by Image-GPT is mapped by the dVAE back to the pixel space $\hat{D}$.
\begin{gather}
    \hat{Z}_D = g_\theta^{GPT}(S^D_{1:N}),\; \hat{D} = g_\theta^{dVAE}(\hat{Z}_D)
\end{gather}
The CET forward pass is shown in Appendix \ref{app:object_centric}.
\subsection{Sequential Prompter:} This (Seq., visualized in Appendix \ref{app:sequential}) is
an ablation over the OCL that replaces CET with an LLM-like sequence prompt obtained from a simple concatenation of slots of images $A$, $B$, and $C$, injected with position embeddings $p_i$.
\begin{gather}
 \mathcal{T}_\alpha(.,.,.) \triangleq  [S_{1:N}^A\;,S_{1:N}^B,\;S_{1:N}^C] + p_i
 \end{gather}
 While the OCL explicitly encodes the entire context into a fixed-length slot sequence via the CET, the ablated agent models a longer concatenated sequence.

\subsection{Patch Learner:} This agent straddles the compositionality extremes of Monolithic Learners and OCLs, and 
uses image patch abstractions. The SA module from OCL is ablated to get a discrete sequence of $4 \times 4$ patch 
latents as the degree of compositional granularity $\mathcal{E}_{\phi}(.) \triangleq f_\phi^{\text{dVAE}}(.)$. A larger image implies longer sequences for this modeling choice.

\section{Experiments}
In this section, we set up experiments to answer (1) whether analogical reasoning enables in-context generalization, (2) does such a generalization require key mechanisms from LLMs, namely compositionality and attention, (3) if 
compositionality is a key ingredient, what is the correct granularity for visual tokens, and (4) what are the limits to 
the generalization.
\label{sec:experiments}

\subsection{Training Setup:}
We trained each agent on the primitive set of tasks across the three benchmarks using Im-Promptu (Section \ref{sec:impromptu}). The SA-based object-centric encoders used in the OCL and Sequential Prompter were pre-trained as a SLATE 
autoencoder, as described by the authors in \cite{singh_SLATE}. On the other hand, the Patch Learner was trained
completely from scratch. The loss metric $\mathcal{L}$ used to train the agents was cross-entropy (CE) loss
between the true latent sequence $Z_D$ and the predicted solution $\hat{Z}_D$ obtained from the Image-GPT, i.e., 
$\mathcal{L}_{impromptu} = CE(Z_D,\hat{Z}_D)$.

In addition to the above loss, the dVAE was trained using the mean-squared error (MSE) loss over the raw pixel space to yield the full loss function $\mathcal{L} =\mathcal{L}_{impromptu} + MSE(D,\hat{D})$. For inference,  answers of the transformers-based agents were sampled from the Image-GPT decoder using top-$k$ nucleus sampling \cite{holtzman2020curious}. Hyperparameters for training and inference have been laid out in Appendix \ref{app:hyperparams}.

\subsection{Primitive Task Extrapolation:}
\label{sec:source_target}
In the first out-of-distribution paradigm, we tested the ability of agents to apply learned rules to different
domain pairs. In order to do this, we held out $20\%$ of  source-target pairs (see Section \ref{sec:benchmark} for
the definition of source and target) from the training primitives and tested the ability of agents to generalize
from learned rules to unseen source-target pairs. For example, the \texttt{object} \texttt{color} property in the
3D Shapes benchmark can take $10$ unique values and $10 \times 9 = 90$ source-target combinations. Thus, an agent was 
trained on only $90 \times 0.8 = 72$ pairs for object-hue primitives. We made sure that each value in the domain set was shown at least once as either the target or the source. 

Fig.~\ref{fig:results_source_target}(a) plots scores of different agents across benchmarks against two key
metrics: (1) MSE (lower is better) that quantitatively compares the construction against the ground truth and 
(2) Fréchet inception distance (FID, lower is better) score to measure the perceptual quality of the composition. We 
make several interesting observations:

\begin{figure}
    \centering
    \includegraphics[width=\textwidth]{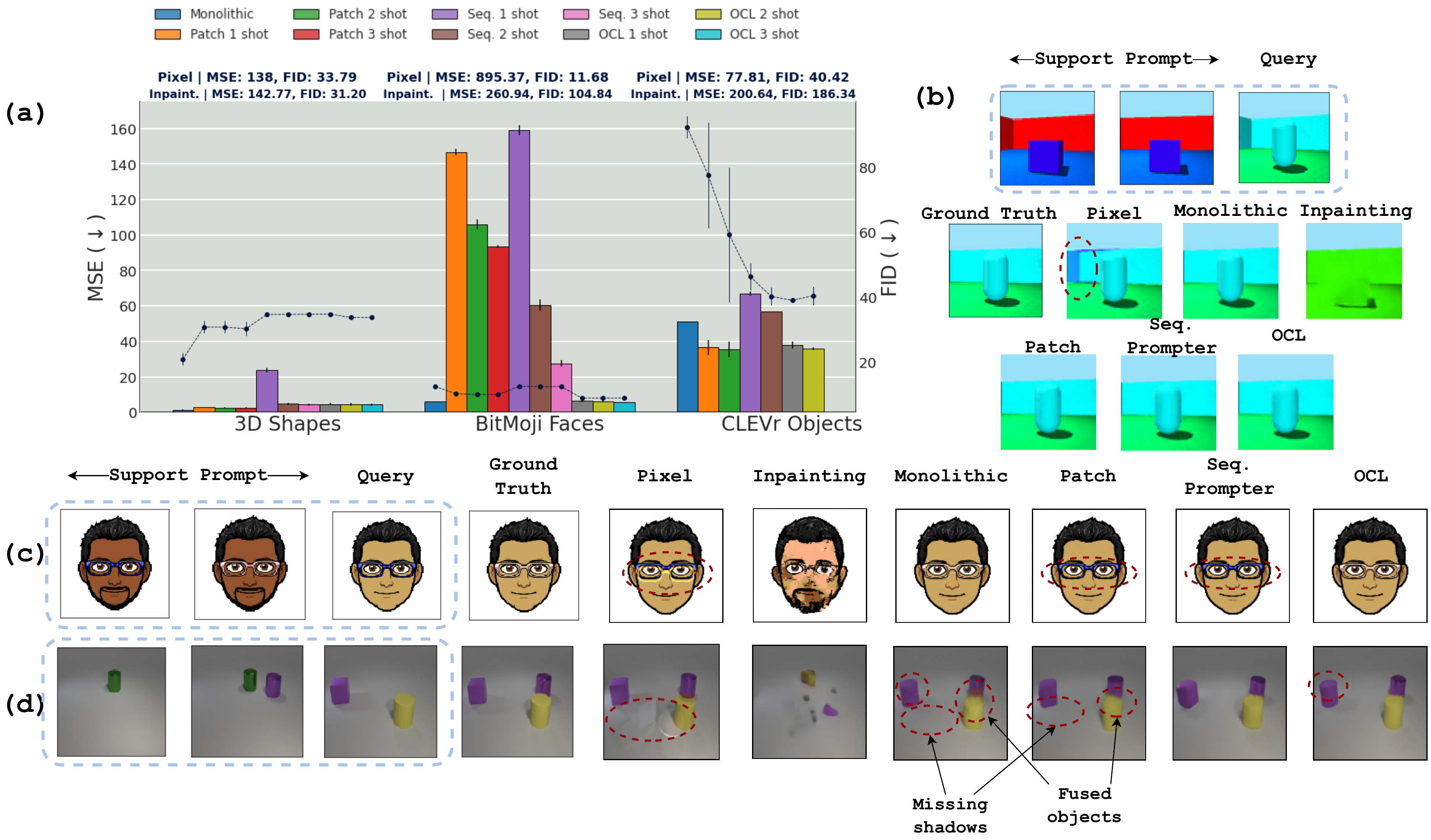}
    \caption{Primitive task extrapolation: (a) Plot showing scores where the left y-axis and the bars represent the 
MSE scores, while the right y-axis and the dotted line denote the FID scores. (b)-(d) Comparison of agent solutions 
with the dotted red circle representing anomalies with respect to the ground truth.} 
    \label{fig:results_source_target}
\end{figure}

\textbf{(R1.1) Simple pixel manipulation produces implausible outputs.}  
The baseline has poor MSE scores ($\sim$ 138 for 3D Shapes, $\sim$ 895 for BitMoji) with clearly apparent
artifacts in the output. While linear pixel transformations are sufficient when an object is independent of the
remaining entities, it is unable to model complex dependencies (see Fig.~\ref{fig:results_source_target}(b),(d) under 
`Pixel').

\textbf{(R1.2) Inpainting model struggles to generalize beyond i.i.d. inputs.} We observed that the Inpainting model is able to cogently complete analogies on the i.i.d. validation set (see Appendix 
\ref{app:inpainting_train} for outputs). However, when tested on extrapolated primitives, it only fills in the high-level structure (see Fig. \ref{fig:results_source_target} (b), (c)) and, in the case of more complex inputs, produces entirely incoherent outputs (Fig. \ref{fig:results_source_target} (d)). We posit that a top-down method like inpainting pre-ordains larger datasets for improved generalization and, as such, suffers from sample inefficiency.  

\textbf{(R1.3) Monolithic Learners are reasonably effective at generalizing from learned rules on structured stimuli.} 
These agents have the lowest MSE scores on 3D Shapes and BitMoji Faces, both spatially consistent datasets.  However, 
the output deteriorated over CLEVr Objects where the latent scene configuration is diverse. The edges of the generated
shapes and lighting details were significantly blurred, and occluded objects were often fused together (FID $\sim$
92, worse than Pixel baseline, Fig.~\ref{fig:results_source_target}(d) under `Monolithic'). 

\textbf{(R1.4) Patch Learner and Sequential Prompter benefit from additional examples.} For these agents, the
addition of context examples was markedly apparent from the significant MSE drops for $>$ 1 shot (see 
Fig.~\ref{fig:results_source_target}(a)). We posit that since these models learn from longer contexts, additional examples are particularly effective in regularizing the output sequence manifold of the Image-GPT. 
\begin{figure}[!hbt]
    \centering
    \includegraphics[width=\textwidth]{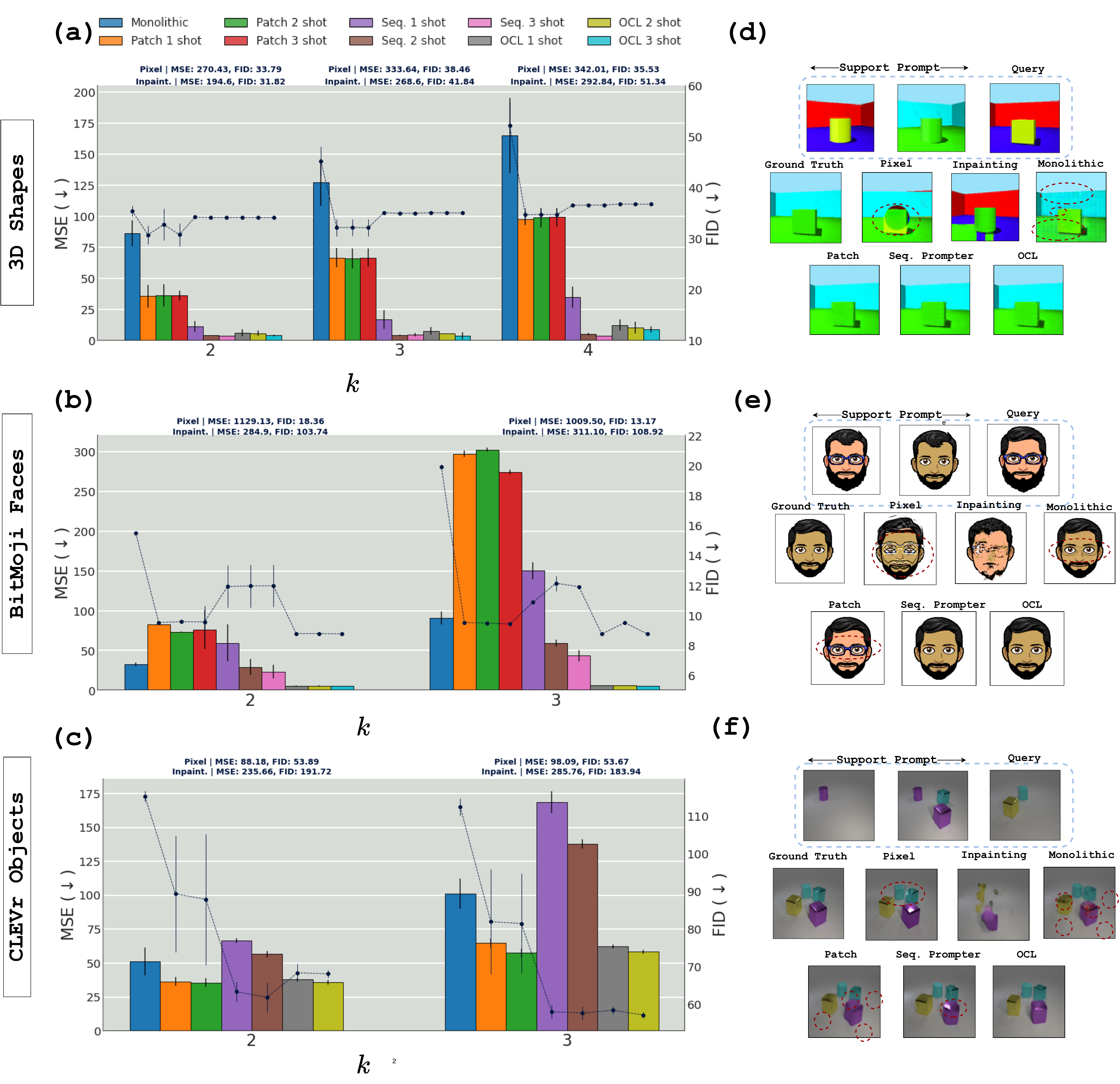}
    \caption{Composite task extrapolation: (a)-(c) Plots showing scores across benchmarks where the left y-axis and
the bars represent the MSE scores, while the right y-axis and the dotted line denote the FID scores. $k$ denotes the level of compositeness. (d)-(f)
Comparison of agent solutions with the dotted red circle representing anomalies with respect to the ground truth. } 
    \label{fig:results_composite}
\end{figure}

\subsection{Composite Task Extrapolation:} 
\label{sec:composite}
In this generalization paradigm, we probed the compositional generalization capabilities of the agents. A 
$k$-composite task ($k\geq 2$, see Section \ref{sec:benchmark}) is an unordered composite function of $k$ primitives.  The value of $k$ is varied across each benchmark, where higher values of $k$ yield visually complex contexts. To solve such a task, the agent must identify the underlying primitive relations and combinatorially modify the query.

In Fig.~\ref{fig:results_composite}(a), we note (R1.1) and (R1.3) from the previous setup in this extrapolation paradigm as well, 
 but we make additional key observations next:

\textbf{(R2.1) The effect of object-centric biases is strong.} This is indicated by the widening MSE gap between
the OCL and monolithic agent for increasing values of $k$ across all three benchmarks (see 
Fig.~\ref{fig:results_composite}(a)-(c)). Moreover, the OCL and the Sequential Prompter generate outputs with consistent perceptual quality (lower FID scores).

\textbf{(R2.2) Monolithic and Patch Learners apply shortcuts.} The monolithic agent tends to produce a
Frankenstein-like image, as if a superposition of context images (see Fig.~\ref{fig:results_composite}(d)-(f)
under `Monolithic'). In the case of the CLEVr Objects dataset, both the Monolithic and Patch Learners often simply
filled in `blobs' agnostic to the color and shape of objects (Fig.~\ref{fig:results_composite}(f), Appendix 
\ref{app:composite_task}), with missing shadows leading to lower MSE scores but poor visual quality. 

\subsection{Analysis:} Next, we discuss key elements required for in-context composition.

\textbf{(A1) In-context generalization critically hinges on compositional granularity.} As evident from (R1.2),
monolithic representations suffice for reliably learning primitive composition rules over simple contexts.
However, they suffer from fidelity issues when the visual space becomes more complex. Beyond primitive
generalization, patch abstractions are effective when the underlying space has object-patch equivalence, as evident 
from the CLEVr Objects dataset 
(Fig.~\ref{fig:results_composite}(c)). However, even with object-patch equivalence, the Patch Learner omits the 
inter-object dependencies (e.g., object occlusions and shadows). It further fails to learn complex and diverse
objects spanning multiple patches reliably (does poorly on the BitMoji Faces dataset with diverse facial
components). Slot-based compositionality with the CET not only enables the generalization of the learned primitives 
but also enables combinatorial generalization to composite tasks that require abstracting out the composition rule over each entity in isolation as well as composing the resultant interplay between them to generate a globally consistent output.

\textbf{(A2) Encoding the context via cross-attention leads to sample efficiency.} Longer slot sequences in the 
Sequential Prompter require multi-shot contexts across all benchmarks. The sample inefficiency was particularly
observed on the CLEVr Objects dataset that uses six slots and, hence, much longer sequences. Cross-attention via 
CET is essential for implicit inference, enabling the learner to encode specific objects from the context.  




\subsection{`Counterfactual Prompt Engineering' with Im-Promptu:}

\begin{figure}[!htb]
  \centering
  \includegraphics[width=\linewidth]{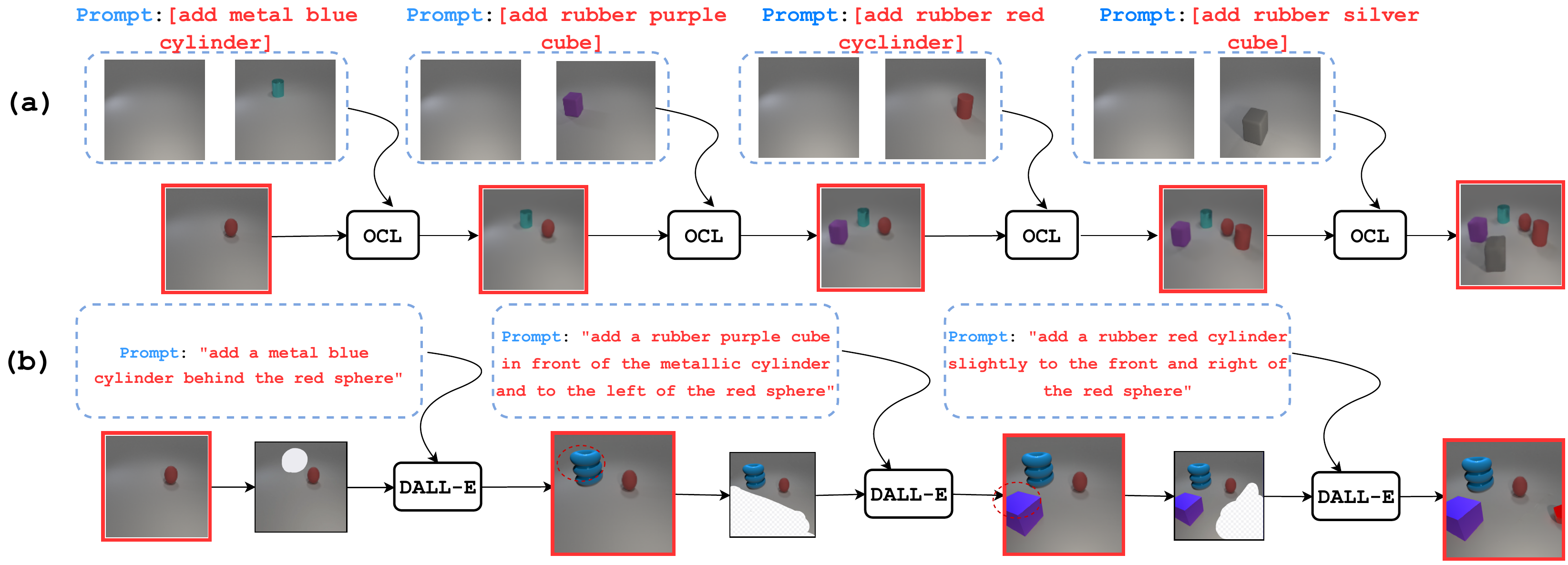}
  \caption{Scene creation from object components: (a) OCL trained via Im-Promptu is used to generate a scene of objects. 
Image-Prompts engineered using existing assets reliably generate the scene object by object. (b) The same scene is 
generated via language grounding and in-painting features of DALL-E \cite{dalle}. Object properties are distorted and 
specifying the location of objects is tedious via language.}
  \label{fig:impromptu}
\end{figure}

\label{sec:prompt}
 LLMs have spurred interest in several priming techniques \cite{liu2021pretrain, willison2023} that can `engineer'
the model into following user instructions. While this natural-language-based programming language provides an
intuitive interface for tasks that have a fundamental degree of language abstraction (writing code, summarization,
symbolic manipulation, etc.), it is not \textit{a priori} obvious that textual descriptions should also scaffold
image generators. Coming up with natural-language prompts for complex visual entities is often tedious and
unintuitive. Instead, explaining an image is naturally easier from its entities and composition over
them\footnote{``No one is an artist unless he carries his picture in his head before painting it, and is sure of
his method and composition'' -- Claude Monet}. To this end, Im-Promptu provides an avenue for generating
counterfactuals via image prompts. In Fig.~\ref{fig:impromptu}(a) we demonstrate the creation of a scene from its
object components via the OCL. The user provides the object components in the form of image prompts constructed from existing images. As a point of comparison, in Fig.~\ref{fig:impromptu}(b), using image-inpainting and natural language prompts with DALL-E \cite{dalle} yields an unreliable output with distorted objects. More practically, 
the OCL reduces human effort significantly in rendering images with desired properties by using existing visual assets.

\section{Conclusion}
This work investigated whether analogy-solving can enable in-context compositional generalization over visual
entities.  We first designed a new test suite. We probed if attention and compositionality, key
ingredients in LLMs, also play a crucial role in the visual domain. We transported these ingredients from language
to model several agents. Our experiments showed that non-compositional agents do not generalize well beyond primitive
task extrapolation. Even with compositionality baked in, the performance of patch-based learners depends on 
idiosyncrasies in the visual structure. However, we found that object-centric biases consistently facilitate an implicit
understanding of composition rules and generate outputs with global semantic consistency. Our ablation of the CET 
suggested that cross-attention plays a crucial role, just like language models, which posits the importance of analogy and 
attention in understanding the underpinnings of in-context learning. 
Future research challenges include collecting real-world primitives from photo-realistic 
graphic engines or human labels, improving object-centric inductive biases for real-life entities, and designing 
better prompting techniques. We hope that our work will spur further research on visual in-context learning.


\begin{ack}
We would like to thank Sreejan Kumar for helpful discussions and insights throughout the course of this project. 
The experiments reported in this article were performed on the computational resources managed and supported by 
Princeton Research Computing at Princeton University. This project was funded by NSF under Grant No. CCF-2203399 and ONR Grant No. N00014-18-1-2873.
\end{ack}

\bibliography{references}
\bibliographystyle{unsrt}
\nocite{*}
\newpage
\appendix
\appendixpage

\section{Benchmark Details}
\begin{figure}[H]
    \centering
    \includegraphics[width=\textwidth]{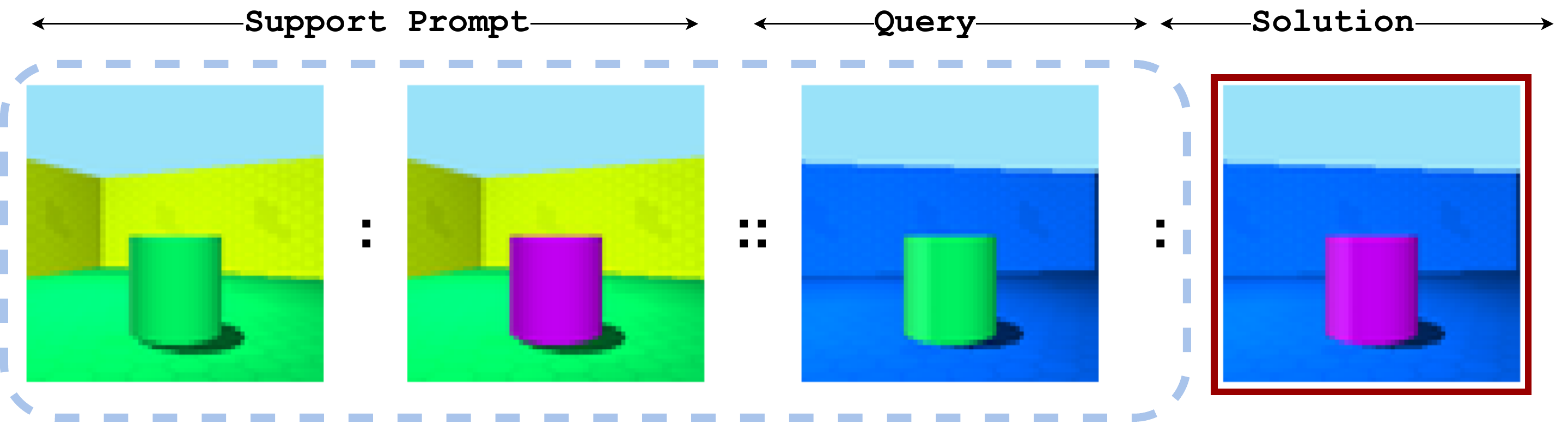}
    \caption{\textbf{3D Shapes} $\rightarrow$  Modify \texttt{object color} from {\color{red}source value} to {\color{blue}target value}.}
    \label{fig:object_hue}
\end{figure}

\begin{figure}[H]
    \centering
    \includegraphics[width=\textwidth]{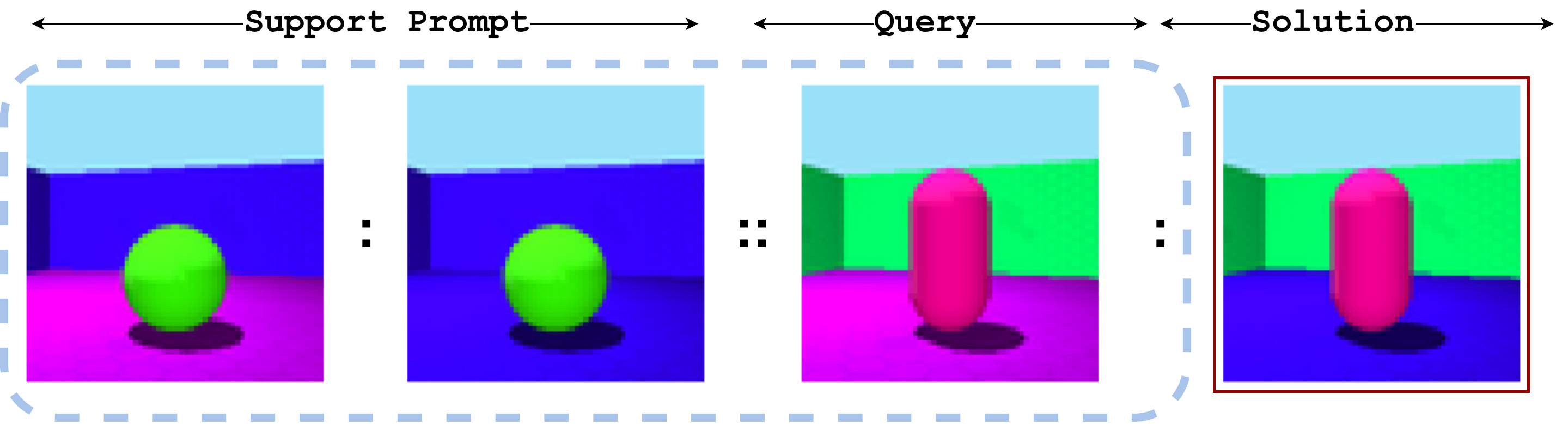}
    \caption{\textbf{3D Shapes} $\rightarrow$  Modify \texttt{floor color} from {\color{red}source value} to {\color{blue}target value}.}
    \label{fig:floor_hue}
\end{figure}

\begin{figure}[H]
    \centering
    \includegraphics[width=\textwidth]{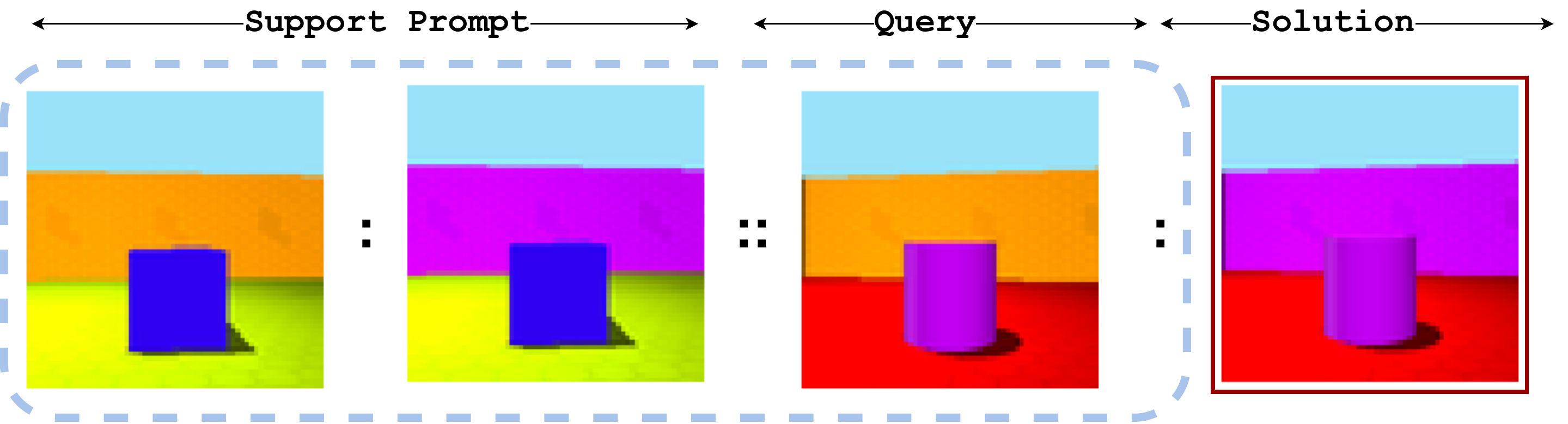}
    \caption{\textbf{3D Shapes} $\rightarrow$  Modify \texttt{wall color} from {\color{red}source value} to {\color{blue}target value}.}
    \label{fig:wall_hue}
\end{figure}

\begin{figure}[H]
    \centering
    \includegraphics[width=\textwidth]{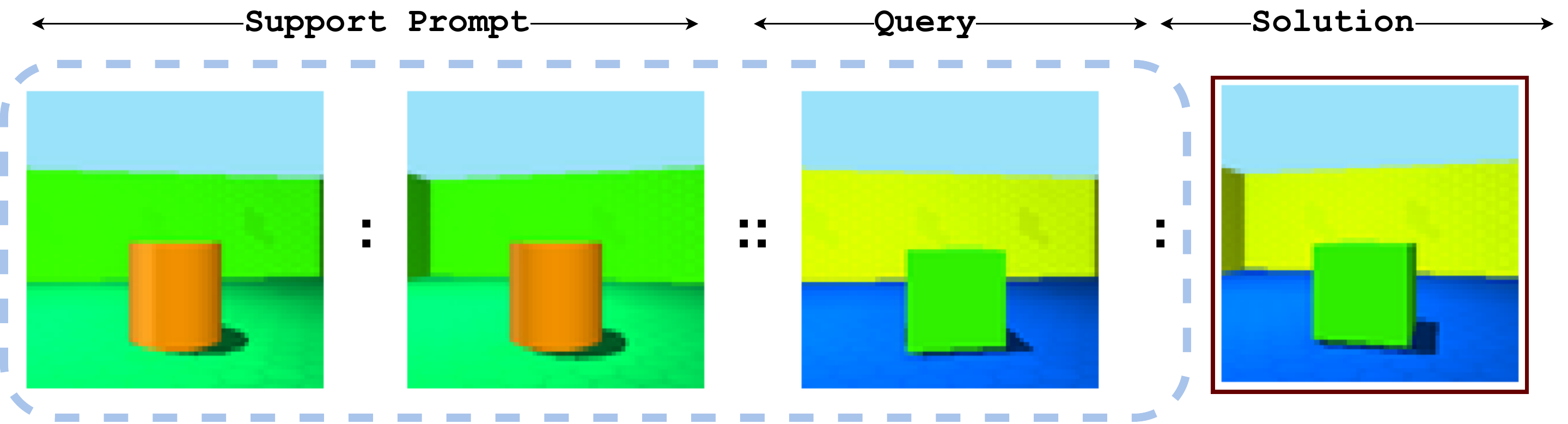}
    \caption{\textbf{3D Shapes} $\rightarrow$  Modify \texttt{orientation} from {\color{red}source value} to {\color{blue}target value}.}
    \label{fig:orientation}
\end{figure}

\begin{figure}[H]
    \centering
    \includegraphics[width=\textwidth]{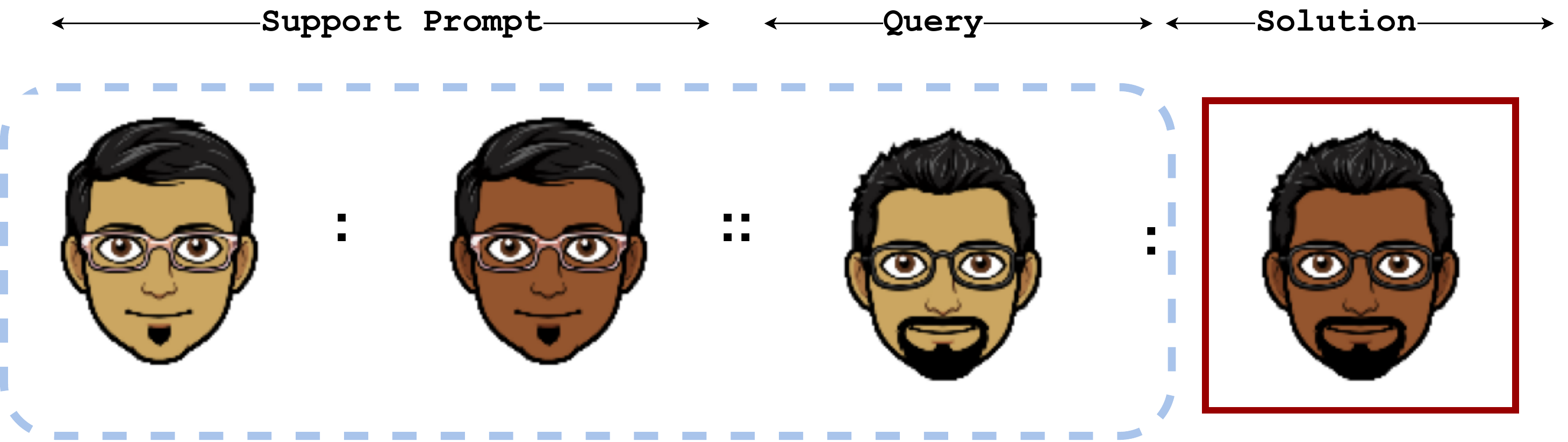}
    \caption{\textbf{BitMoji Faces} $\rightarrow$  Modify \texttt{facial skintone} from {\color{red}source value} to {\color{blue}target value}.}
    \label{fig:skin_tone}
\end{figure}

\begin{figure}[H]
    \centering
    \includegraphics[width=\textwidth]{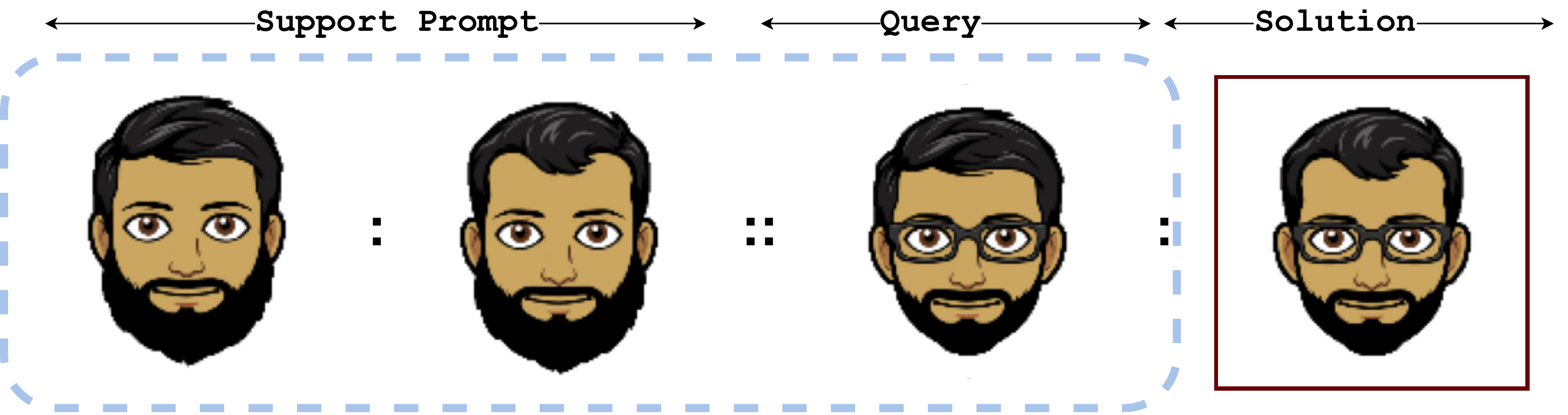}
    \caption{\textbf{BitMoji Faces} $\rightarrow$  Modify \texttt{hair style} from {\color{red}source value} to {\color{blue}target value}.}
    \label{fig:hair_style}
\end{figure}

\begin{figure}[H]
    \centering
    \includegraphics[width=\textwidth]{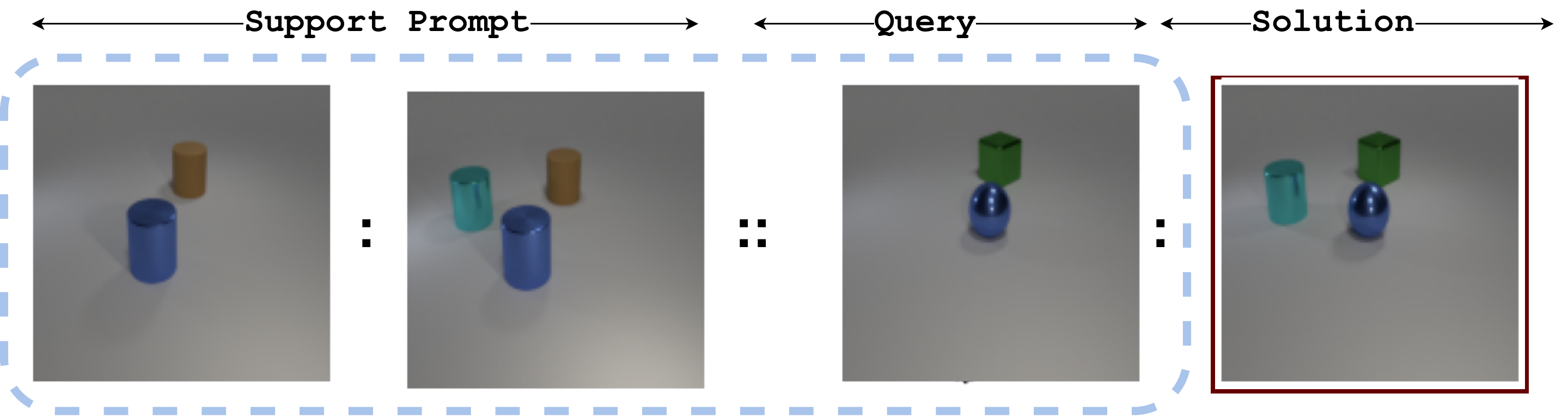}
    \caption{\textbf{CLEVr} $\rightarrow$ Add object to position {\color{blue}$p$}.}
    \label{fig:add_object}
\end{figure}

\begin{figure}[H]
    \centering
    \includegraphics[width=\textwidth]{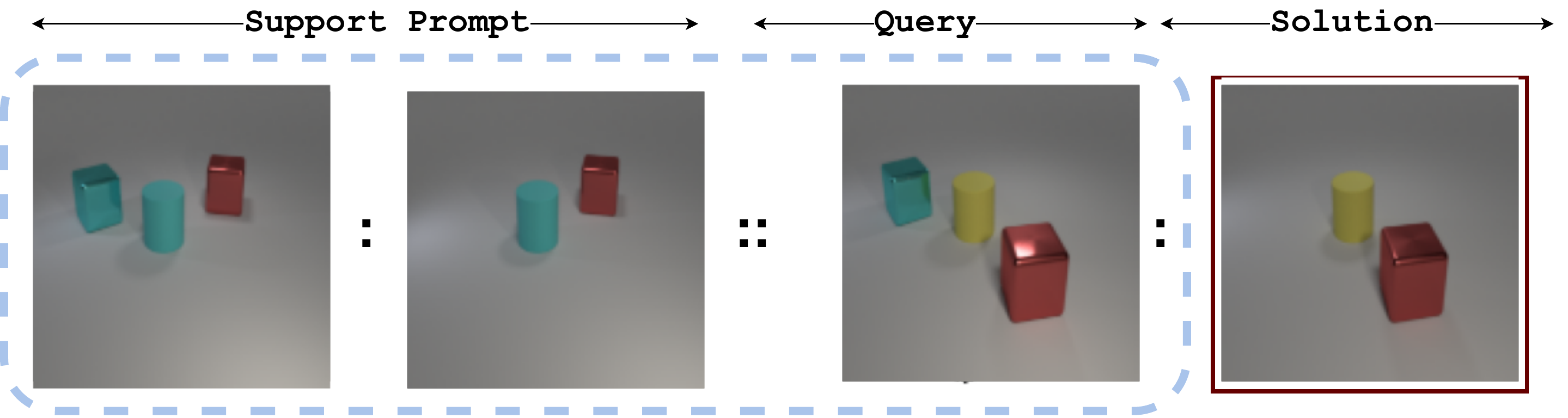}
    \caption{\textbf{CLEVr} $\rightarrow$ Delete object from position {\color{blue}$p$}.}
    \label{fig:delete_object}
\end{figure}

The full set of primitive tasks is summarized in Table \ref{tab:benchmark} and the statistics detailed in Table
\ref{tab:stats}. Primitives can be individually visualized in Figs.~\ref{fig:object_hue}-\ref{fig:delete_object}. Each primitive figure shows an analogy where the underlying relation is over a particular composable property of the scene/image. 
\label{app:benchmarks}


\begin{table}[!htb]
  \centering
  \begin{tabular}{llc}
    \toprule             
    Dataset  & Primitive Task Set & Domain Size  \\
    \midrule
    \multirow{4}{*}{3D Shapes} &  (1) Modify \texttt{object color} from {\color{red}source value} to {\color{blue}target value} &  10  \\
    & (2) Modify \texttt{floor color} from {\color{red}source value} to {\color{blue}target value} & 10\\
    & (3) Modify \texttt{wall color} from {\color{red}source value} to {\color{blue}target value} & 10\\
    & (4) Modify \texttt{scene orientation} from {\color{red}source value} to {\color{blue}target value} & 15 \\
\midrule
\multirow{4}{*}{BitMoji Faces} & \t (1) Modify \texttt{skintone} from {\color{red}source value} to {\color{blue}target value} & 3 \\
    & (2) Modify \texttt{hair style} from {\color{red}source value} to {\color{blue}target value} & 10\\
    & (3) Modify \texttt{facial hair} from {\color{red}source value} to {\color{blue}target value} & 5\\
    & (4) Modify \texttt{eyewear} from {\color{red}source value} to {\color{blue}target value} & 5\\
\midrule
\multirow{2}{*}{CLEVr Objects} & (1) Add object to the scene at position {\color{blue} $p$} & $1000$ \\
& (2) Delete object from the scene at position {\color{blue} $p$} & $1000$\\
\bottomrule \\
  \end{tabular}
    \caption{Summary of primitive tasks across all benchmarks}
  \label{tab:benchmark}
\end{table}

\begin{table}[!htb]

  \centering
  \begin{tabular}{lcccc}
    \toprule             
    Dataset  & \specialcell{Examples per \\ task} & \specialcell{\#primitive \\ tasks} & \specialcell{\#primitive \\ extrapolation tasks} & \specialcell{\#$k$-composite \\ tasks}\\
    \midrule
3D Shapes & 3 & 80000 & 1000 & 1000 $\forall k \in \{2,3,4\}$ \\
\midrule
BitMoji Faces & 3 & 80000 & 1000 & 1000 $\forall k \in \{2,3\}$ \\
\midrule 
CLEVr Objects & 2 & 55000 & 1000   & 200 $\forall k \in \{2,3,4\}$  \\
\bottomrule \\
  \end{tabular}
    \caption{Benchmark statistics}
      \label{tab:stats}
\end{table}

\begin{figure}[H]
    \centering
    \includegraphics[width=\textwidth]{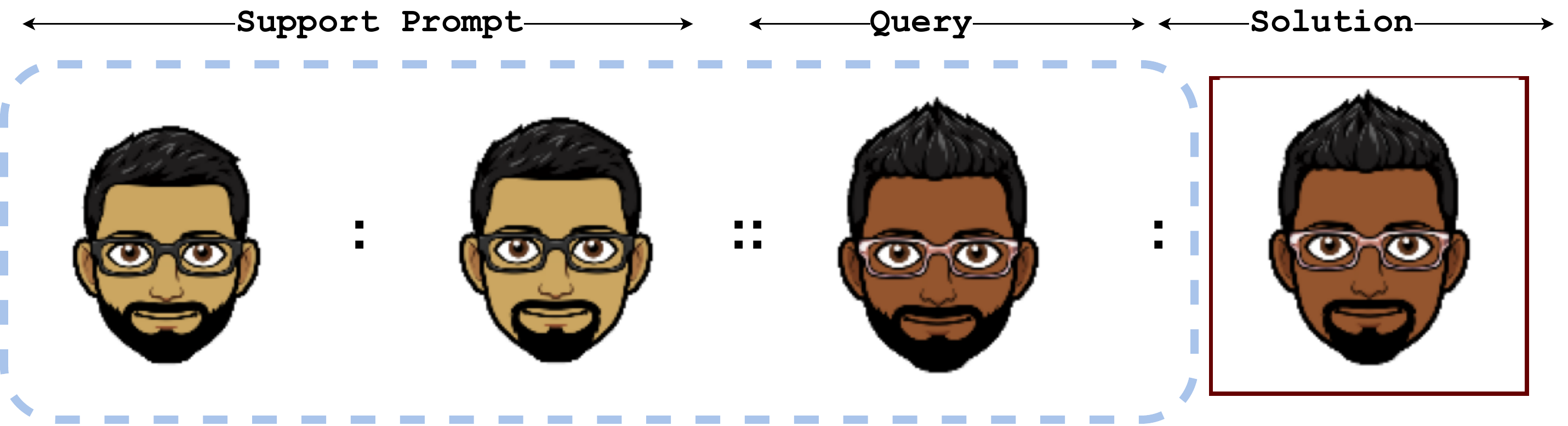}
    \caption{\textbf{BitMoji Faces} $\rightarrow$  Modify \texttt{facial hair} from {\color{red}source value} to {\color{blue}target value}.}
    \label{fig:facial_hair}
\end{figure}

\begin{figure}[H]
    \centering
    \includegraphics[width=\textwidth]{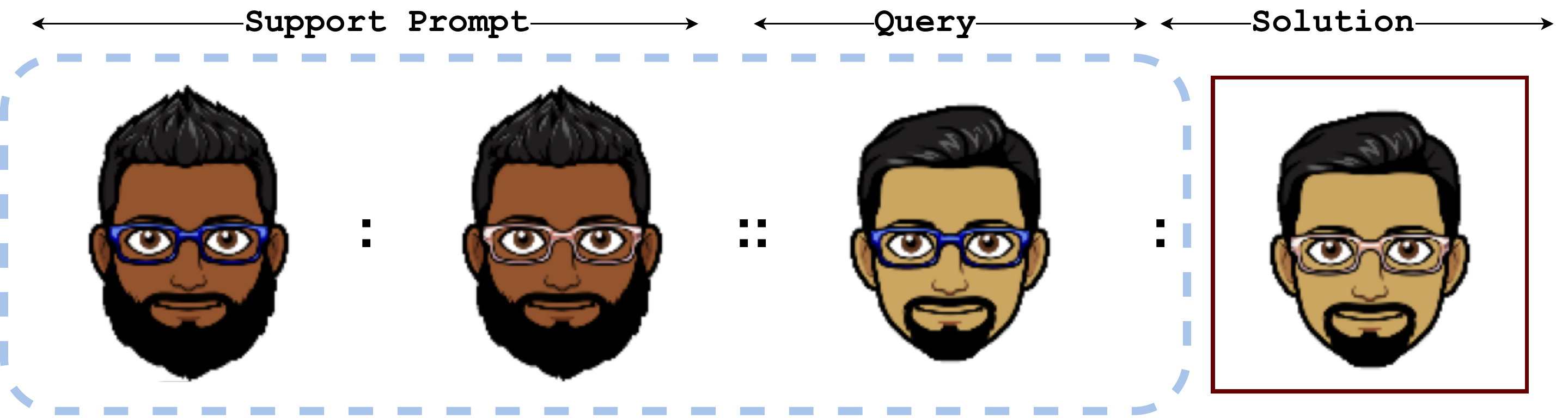}
    \caption{\textbf{BitMoji Faces} $\rightarrow$  Modify \texttt{eyewear} from {\color{red}source value} to {\color{blue}target value}.}
    \label{fig:facial_hair}
\end{figure}

\section{LLMs as In-Context Compositional Learners}

\label{app:LLMs}

It is straightforward to see how the aformentioned framework subsumes LLMs. The input space $\mathcal{X}$ is natural 
language and the compositional space $\mathcal{C}$ contains the dictionary of resultant tokens obtained via a
tokenizer $\mathcal{E}_\phi(.)$. The relation class $\mathcal{R}$ in this context is the set of underlying tasks
spanning numerical reasoning, reading comprehension, sentiment analysis, and question answering, to name a few.
The core Transformer model forms the executor $\mathcal{T}_\alpha(.,.,.)$ that simply concatenates $A,\;B$, and $C$ into a sequence prompt to compose the output $D$ autoregressively. A de-tokenizer $\mathcal{D}_\theta(.)$ constructs natural language outputs from the generated tokens.

\section{Learning Agents}

\subsection{Monolithic:} 
\label{app:monolithic}
The encoder $e_\phi (.)$ and decoder $d_\theta(.)$ architectures are convolutional and deconvolutional networks
with four layers initialized as shown in Tables \ref{tab:encoder_monolithic} and \ref{tab:decoder_monolithic}, respectively.

\begin{table}[H]
    \centering
    \begin{tabular}{c c c c c}
    \toprule
       Layer & Stride & Activation & Channels & Weight Matrix \\
    \midrule
        Conv $3 \times 3$ & 2 & ReLU & 192 & - \\
        Conv $3 \times 3$ & 2 & ReLU & 192 & - \\
        Conv $3 \times 3$ & 2 & ReLU & 192 & - \\
        Conv $3 \times 3$ & 2 & ReLU & 192  & -\\
        Fully Connected & - & ReLU & - & $(192 \times \text{Image Size}/16 \times  \text{Image Size}/16) \times 192$  \\
        Fully Connected & - & - & - & $192 \times 192$ \\
        \bottomrule \\
    \end{tabular}
    \caption{Encoder $e_\phi (.)$ of the monolithic agent}
    \label{tab:encoder_monolithic}
\end{table}

\begin{table}[H]
    \centering
    \begin{tabular}{c c c c c}
    \toprule
       Layer & Stride & Activation & Channels & Weight Matrix \\
    \midrule
        Fully Connected & - & - & - & $192 \times 192$ \\
        Fully Connected & - & ReLU & - & $192  \times (192 \times \text{Image Size}/16 \times  \text{Image Size}/16) $  \\
        Conv Transpose $3 \times 3$ & 2 & ReLU & 192 & - \\
        Conv Transpose $3 \times 3$ & 2 & ReLU & 192 & - \\
        Conv Transpose $3 \times 3$ & 2 & ReLU & 192 & - \\
        Conv Transpose $3 \times 3$ & 2 & ReLU & 3  & -\\
        \bottomrule \\
    \end{tabular}
    \caption{Decoder $d_\theta(.)$ of the monolithic agent}
    \label{tab:decoder_monolithic}
\end{table}

The inference network $f_\alpha(.)$ and executor network $h_\beta(.)$ are feed-forward networks, as parameterized 
in Table \ref{tab:inference_monolithic}.

\begin{table}[H]
    \centering
    \begin{tabular}{ccc}
    \toprule
    Layer  & Weight Matrix & Activation  \\
    \midrule
    Fully Connected   & 384 $\times$ 192 & ReLU \\
    Fully Connected   & 192 $\times$ 192 & ReLU \\
    Fully Connected   & 192 $\times$ 192 & ReLU \\
    \bottomrule \\
    \end{tabular}
    \caption{The inference network $f_\alpha(.)$ and executor network  $h_\beta(.)$ of the monolithic agent}
    \label{tab:inference_monolithic}
\end{table}

\subsection{Inpainting Model:} 
\begin{figure}[!hbt]
    \centering
    \includegraphics[scale=0.5]{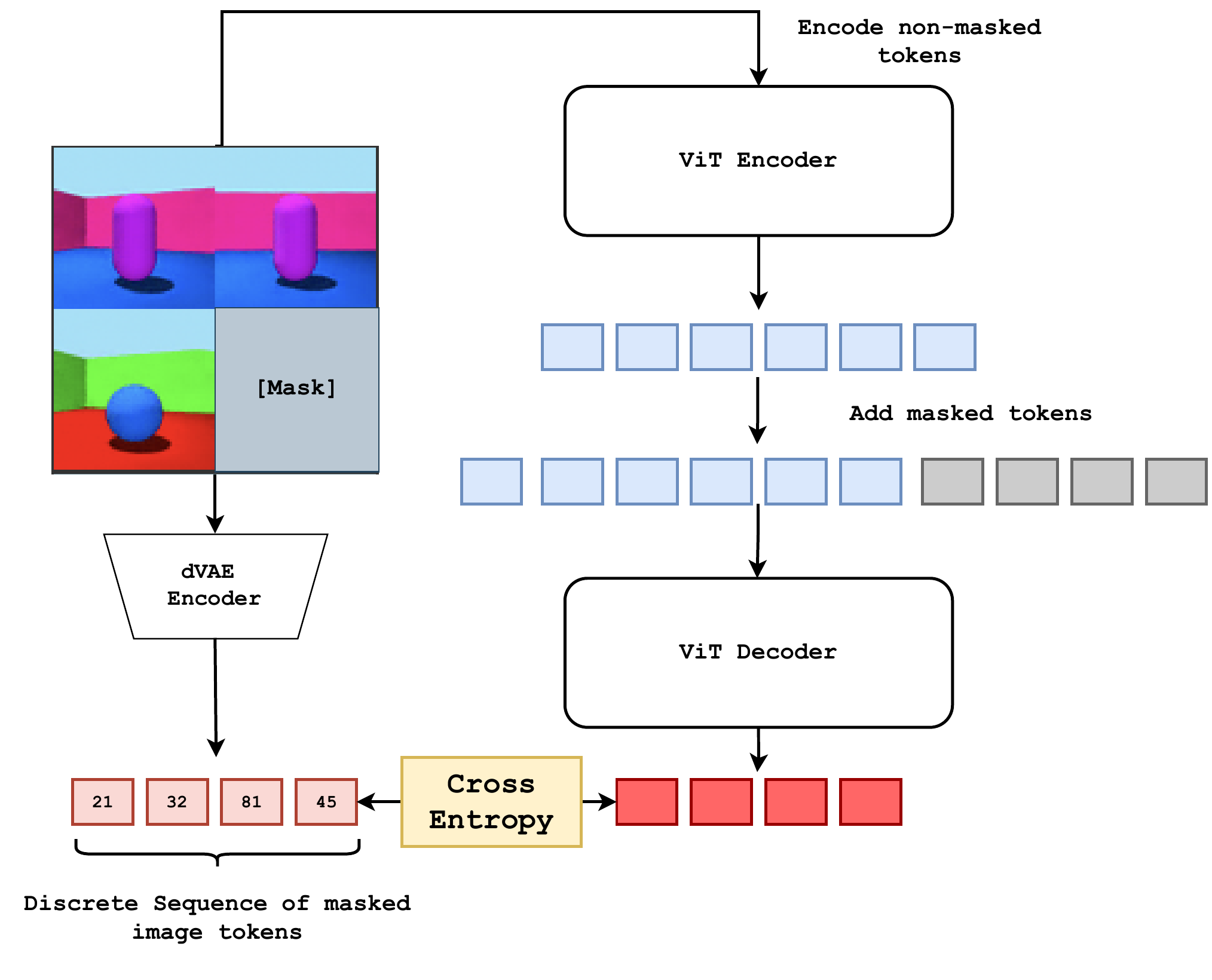}
    \caption{Visualization of the Inpainting model architecture. A dVAE encodes the masked input into a discrete sequence. A ViT \cite{dosovitskiy2021image} encodes the non-masked context patches. Subsequently, masked tokens are added to the encoded patches and decoded via another ViT to predict the discrete sequence. The model is trained end-to-end via the popular Masked Autoencoder (MAE) reconstruction task \cite{he2021masked}}
    \label{fig:inpainting}
\end{figure}

The Inpainting model has been described in Fig. \ref{fig:inpainting}. The model architecture follows 
from the task proposed in \cite{bar2022visual,he2021masked} with the exception of the VQ-GAN replaced by a dVAE and the model trained end-to-end from scratch. Hyperparameters across various test suites have been specified in Table \ref{tab:inpainting_hyperparameters}.
\label{app:inpainting}

\begin{table}[!hbt]
    \centering
    \begin{tabular}{ccccc}
    \toprule
   Module & Hyperparameters &  3D Shapes & BitMoji Faces & CLEVr Objects \\
    \midrule
    
  \multirow{2}{*}{Pixel and Discrete Space}  & Image Size  & 64 & 128 & 128 \\ 
    & Image Tokens & 256 & 1024 & 1024 \\
    \midrule
    \multirow{2}{*}{dVAE}  & Vocabulary Size  & 512 & 1024 & 4096 \\  
     & Tau  & 0.1 & 0.1 & 0.1 \\ 
    \midrule
    \multirow{3}{*}{ViT Encoder/Decoder}  & Num. Layers & 4 & 8 & 8 \\
    & Heads & 4 & 8 & 8 \\
    & Hidden Dims & 192 & 192 & 192 \\
    \midrule
    \end{tabular}
    \caption{Hyperparameters for the Inpainting model.}
    \label{tab:inpainting_hyperparameters}
\end{table}

\subsection{Object-Centric Learner:}
\label{app:object_centric}

The Context Encoder Transformer (CET) is a stack of $L$ blocks, each computing (a) self-attention over inputs followed 
by (b) cross-attention of inputs over context slots. The cross-attention implementation of block $l$  is shown in 
Algorithm \ref{alg:cross_attention}.

 \begin{algorithm}[!hbt]
\caption{Cross-Attention for Block $l$ of CET}
\label{alg:cross_attention}
\begin{algorithmic}

\STATE{\bfseries Require:} $A^{1:N} \in \mathbb{R}^{N \times d}$, $N$ slots of context $A$  \vspace{0.2em} \vspace{0.2em}
\STATE{\bfseries Require:} $B^{1:N} \in \mathbb{R}^{N \times d}$, $N$ slots of context $B$ \vspace{0.2em}
\STATE{\bfseries Require:} $C_{SA,l}^{1:N}  \in \mathbb{R}^{N \times d}$, $N$ slots of input $C$ obtained from the 
self-attention layer of block $l$ \vspace{0.2em}
\STATE{\bfseries Output:} $C_{l}^{1:N}$, Context encoded input slots of block $l$   \vspace{0.2em}
\STATE Get query tokens: $Q^C = MLP_Q(C_{SA,l}^{1:N})$\vspace{0.2em}
\STATE Get keys, values of $A$, $K^A, V^A =MLP_{KV}(A_{l-1}^{1:N})$ \vspace{0.2em}
\STATE Get keys, values of $B$,  $K^B, V^B =MLP_{KV}(B_{l-1}^{1:N})$\vspace{0.2em}
\STATE Concatenate $K=[K^A, K^B] \in  \mathbb{R}^{2N \times d}$ 
\STATE Concatenate $V=[V^A,V^B] \in  \mathbb{R}^{2N \times d} $\vspace{0.2em}
\STATE Compute Cross-Attention Matrix  $Att=\text{softmax}(Q^TK/ \sqrt{d})$
\STATE Compute $C_{l}^{1:N} = Att \times V$

\end{algorithmic}
\end{algorithm}

\subsection{Sequential Prompter:}
\label{app:sequential}

\begin{figure}[!hbt]
    \centering
    \includegraphics[width=\textwidth]{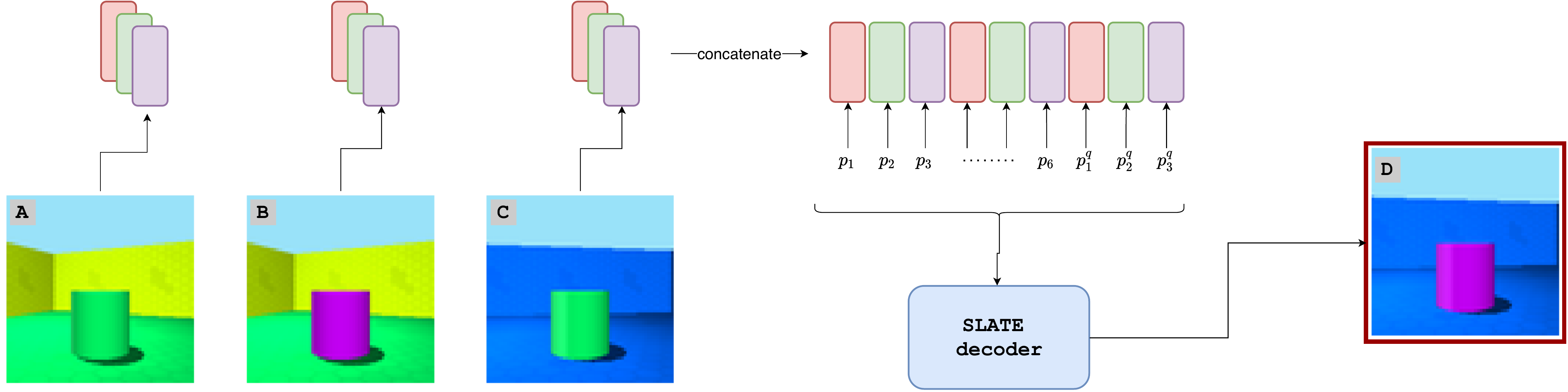}
    \caption{The Sequential Prompter concatenates slots from context prompts and a query together. It further injects position embeddings to create an LLM-like input prompt for the decoder.}
    \label{fig:seq}
\end{figure}

We visualize this learner in Fig.~\ref{fig:seq}.

\newpage

\section{Hyperparameters}
\label{app:hyperparams}

The hyperparameters for OCL are listed in Table \ref{tab:ocl_hyperaparameters}. The ablated architectures follow the same setup without the ablated module. All experiments were performed on NVIDIA A100 GPUs.

\begin{table}[!hbt]
    \centering
    \begin{tabular}{ccccc}
    \toprule
   Module & Hyperparameters &  3D Shapes & BitMoji Faces & CLEVr Objects \\
    \midrule
    
  \multirow{2}{*}{Pixel and Discrete Space}  & Image Size  & 64 & 128 & 128 \\ 
    & Image Tokens & 256 & 1024 & 1024 \\
    \midrule
    \multirow{3}{*}{dVAE}  & Vocabulary Size  & 512 & 1024 & 4096 \\ 
     & LR (no warmup)  & $5\times 10^{-5}$ & $5 \times 10^{-5}$ & $1 \times 10^{-5}$  \\ 
     & Tau  & 0.1 & 0.1 & 0.1 \\ 
    \midrule
    \multirow{3}{*}{CET} & Num. Layers & 4 & 4 & 4\\
    & Heads & 4 & 4 & 4 \\
    & Hidden Dims & 192  &  192 & 192 \\
    \midrule
    \multirow{3}{*}{Image GPT}  & Num. Layers & 4 & 8 & 8 \\
    & Heads & 4 & 8 & 8 \\
    & Hidden Dims & 192 & 192 & 192 \\
    \midrule
    \multirow{5}{*}{Slot Attention} & Num. Slots & 3 & 2 &  6\\
    & Iterations & 3 & 3 & 7\\
    & Slot Heads & 1 & 3 &  1\\
    & Hidden Dims & 192 & 192 & 192\\
    & LR (no warmup) &  $5 \times 10^{-5}$ &  $5 \times 10^{-5}$ & $1 \times 10^{-5}$ \\
    \midrule
     \multirow{5}{*}{Training Setup} & Batch Size & 32 & 24 & 20 \\ 
      & LR Warmup steps  & 15000 & 30000 & 30000 \\ 
       & Peak LR  & $1 \times 10^{-4}$ & $1 \times 10^{-4}$ & $1 \times 10^{-4}$  \\ 
     & Dropout  & 0.1 & 0.1 & 0.1 \\ 
    & Gradient Clipping & 1.0 & 1.0 & 1.0 \\
    \midrule
    \multirow{3}{*}{Sampling Scheme} & Temperature & 0.7 & 0.7 & 0.1 \\
    & Top-$k$ & 8 & 8 & $-$\\
    & Top-$p$ & 0.75 & 0.75 & 0.5 \\
    \midrule
    \multirow{2}{*}{Training Cost} & GPU Usage &  15 GB & 40 GB & 40 GB\\
    & Days & 1 & 3 & 5\\
    \bottomrule \\
    \end{tabular}
    \caption{Hyperparameters for the Object-Centric Learner (OCL) instantiation and  training setup}
    \label{tab:ocl_hyperaparameters}
\end{table}

\newpage
\section{Experiments}

\subsection{Object Slot Emergence:} 
\begin{figure}[!hbt]
    \centering
    \includegraphics[width = \textwidth]{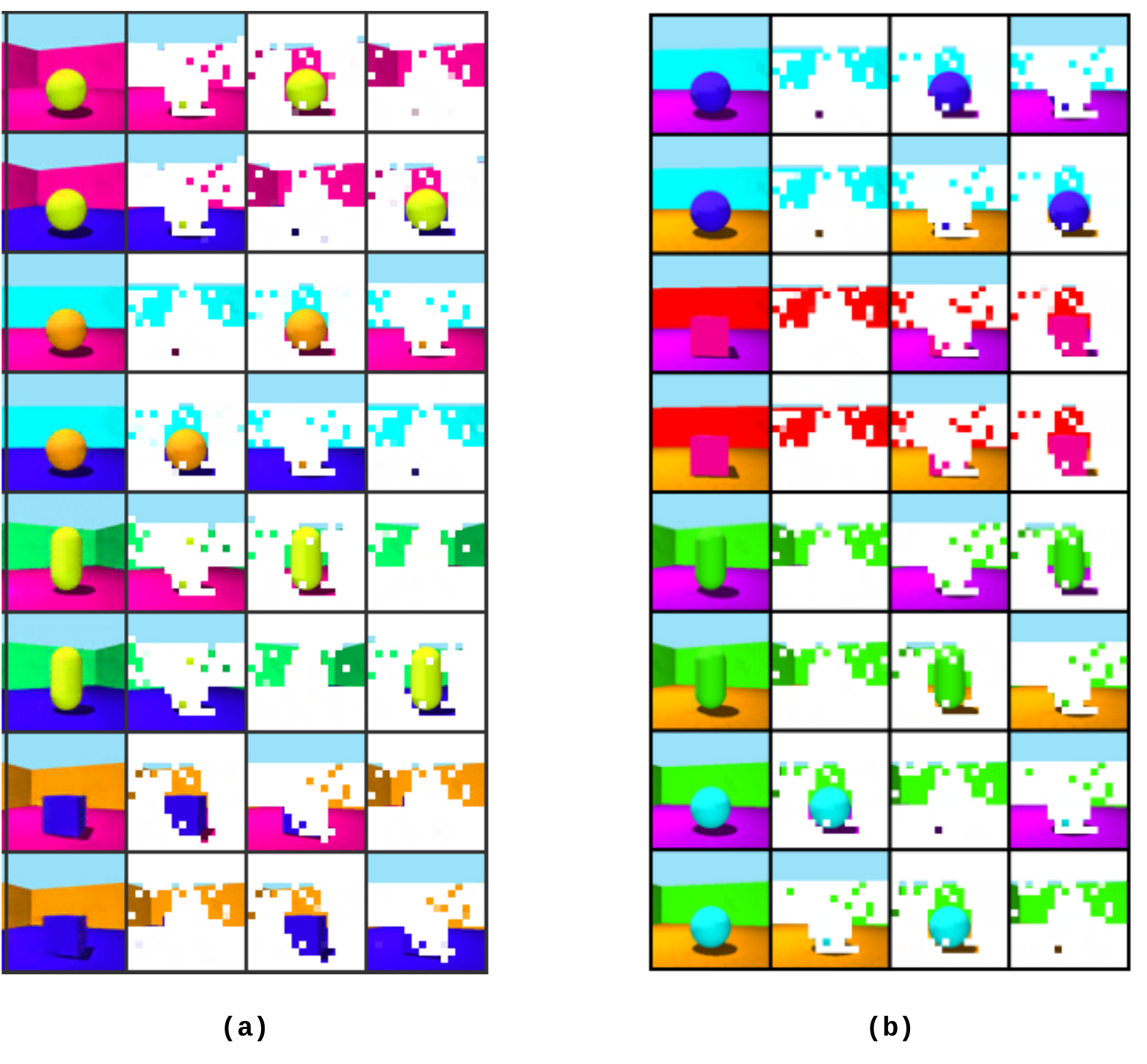}
    \caption{Slots from training OCLs on 3D Shapes: (a) pre-trained slots and (b) slots after Im-Promptu training. 
The structure of slots remains preserved.}
    \label{fig:slot_shapes3d}
    \vspace{1em}
\end{figure}

\begin{figure}[!hbt]
    \centering
    \includegraphics[width = \textwidth]{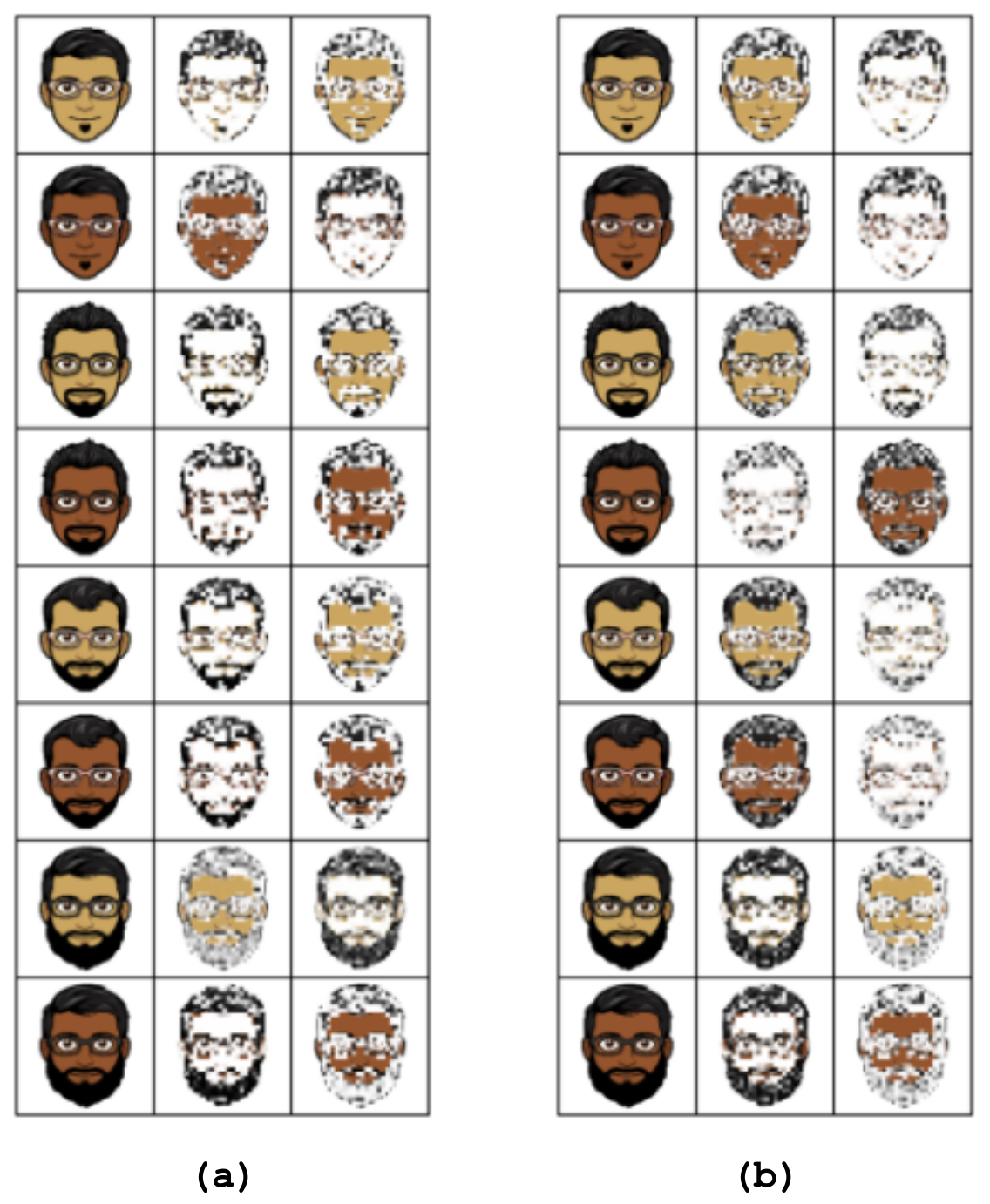}
    \caption{Slots from training OCLs on BitMoji Faces: (a) pre-trained slots and (b) slots after Im-Promptu training. 
There occurs a minor modification in slots where the structure of the eye is merged with the slots carrying the facial 
contents.}
    \label{fig:slot_bitmoji}
\end{figure}

\begin{figure}[!hbt]
    \centering
    \includegraphics[width = \textwidth]{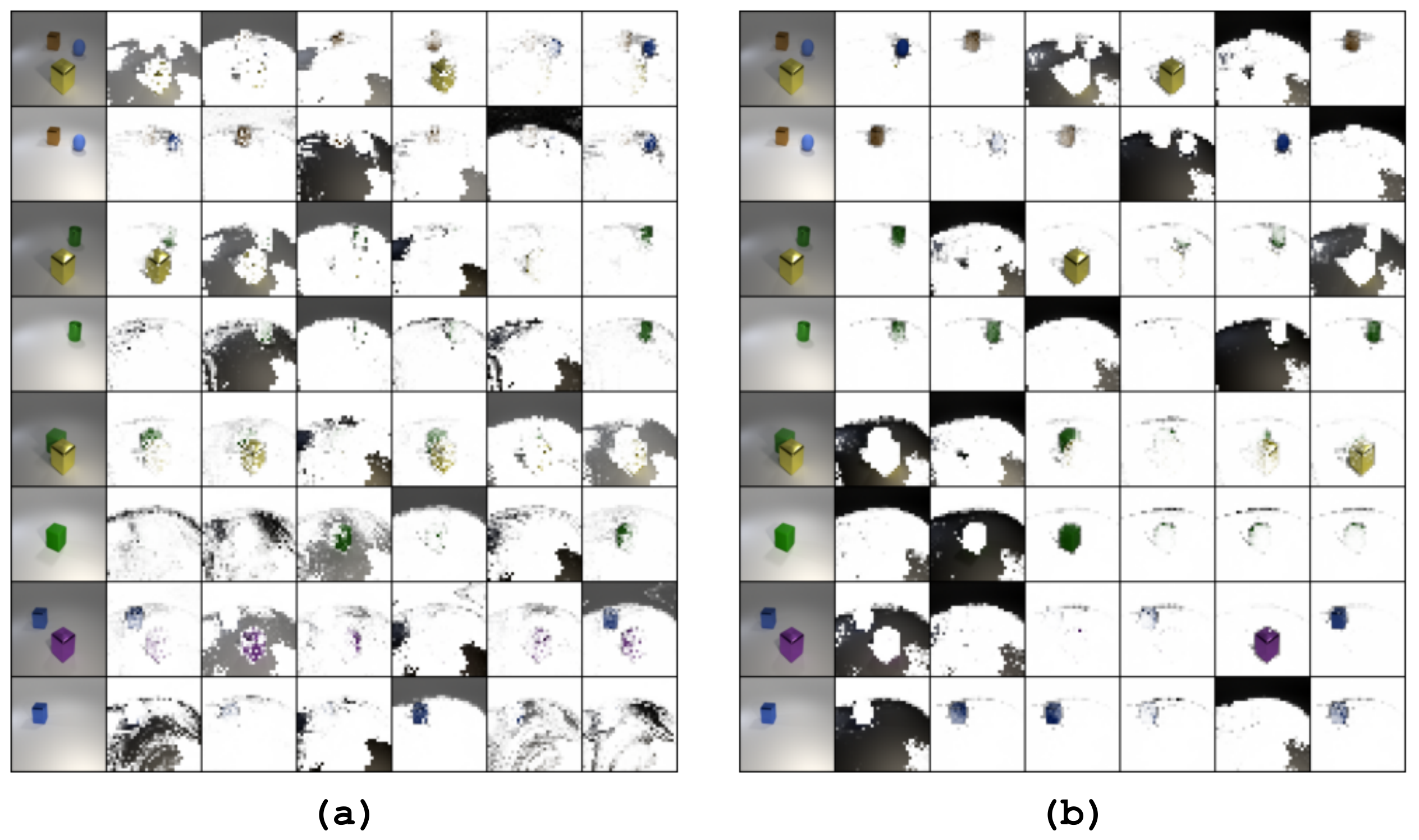}
    \caption{Slots from training OCLs on CLEVr Objects: (a) pre-trained slots and (b) slots after Im-Promptu training. 
We observe that slots become more refined after Im-Promptu training.}
    \vspace{1em}
    \label{fig:slot_clevr}
\end{figure}

Distinct object slots stay preserved in object-centric architectures (OCL and 
Sequential Prompter) while training on primitive in-context learning tasks. We visualize the slot formation in 
Figs.~\ref{fig:slot_shapes3d}, \ref{fig:slot_bitmoji}, and \ref{fig:slot_clevr} for 3D Shapes, BitMoji Faces, and 
CLEVr Objects, respectively.

\subsection{Primitive Task Extrapolation:}
\begin{figure}
    \centering
    \includegraphics[width=\textwidth]{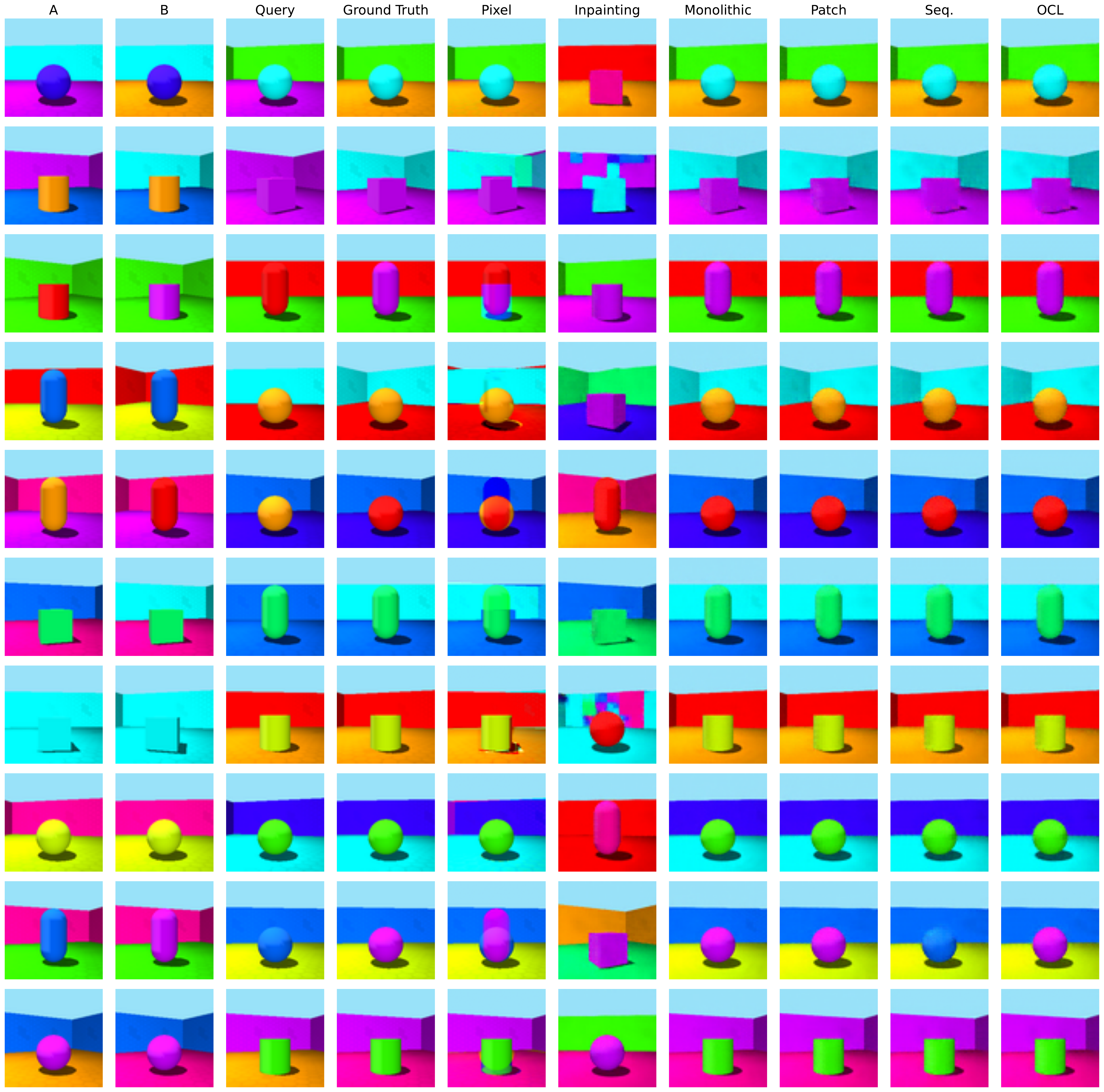}
    \caption{Primitive task extrapolation for the 3D Shapes benchmark.}
    \label{fig:shapes3d_source_target}
\end{figure}

\begin{figure}
    \centering
    \includegraphics[width=\textwidth]{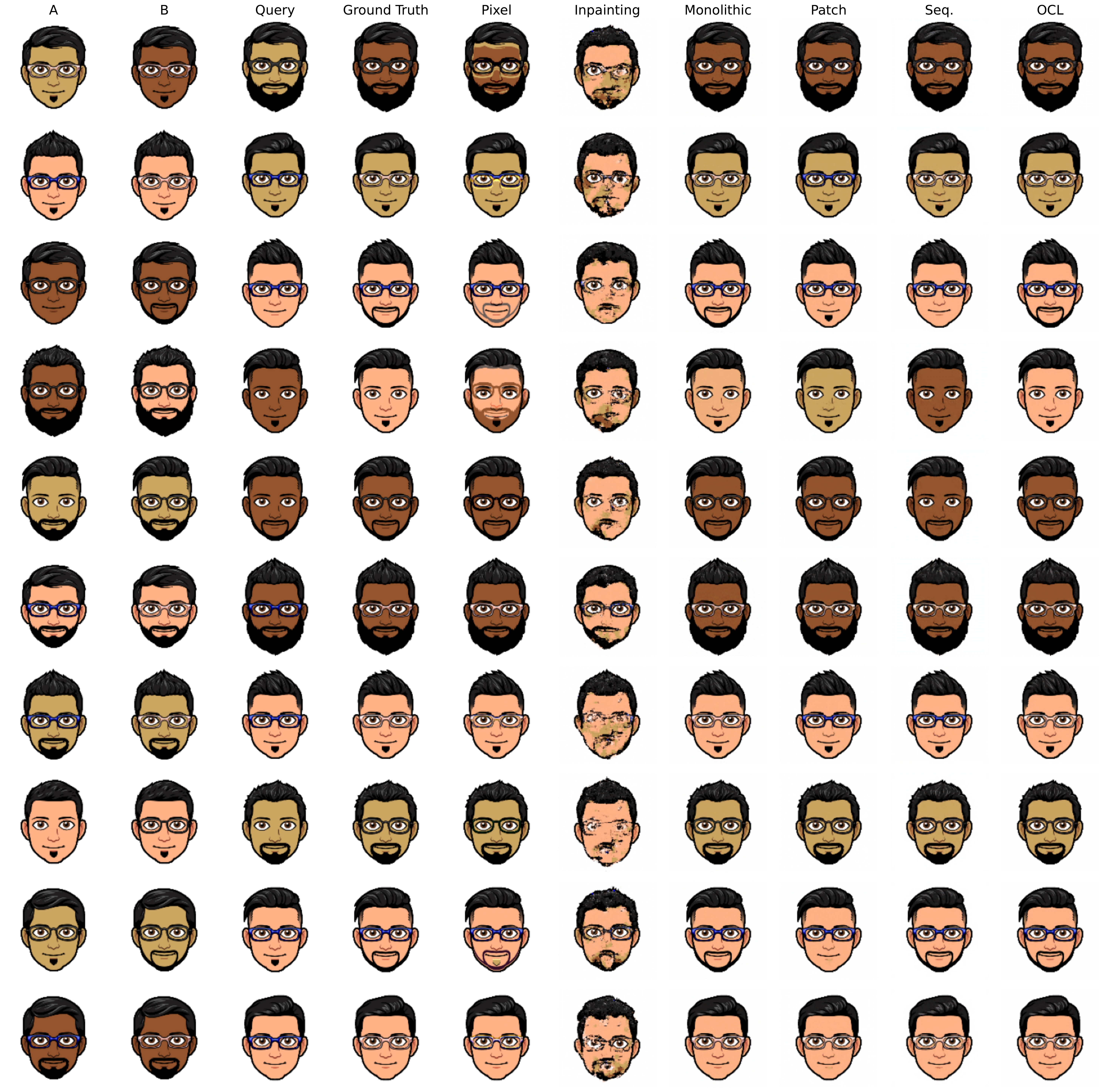}
    \caption{Primitive task extrapolation for the BitMoji Faces benchmark.}
    \label{fig:bitmoji_source_target}
\end{figure}

\begin{figure}
    \centering
    \includegraphics[width=\textwidth]{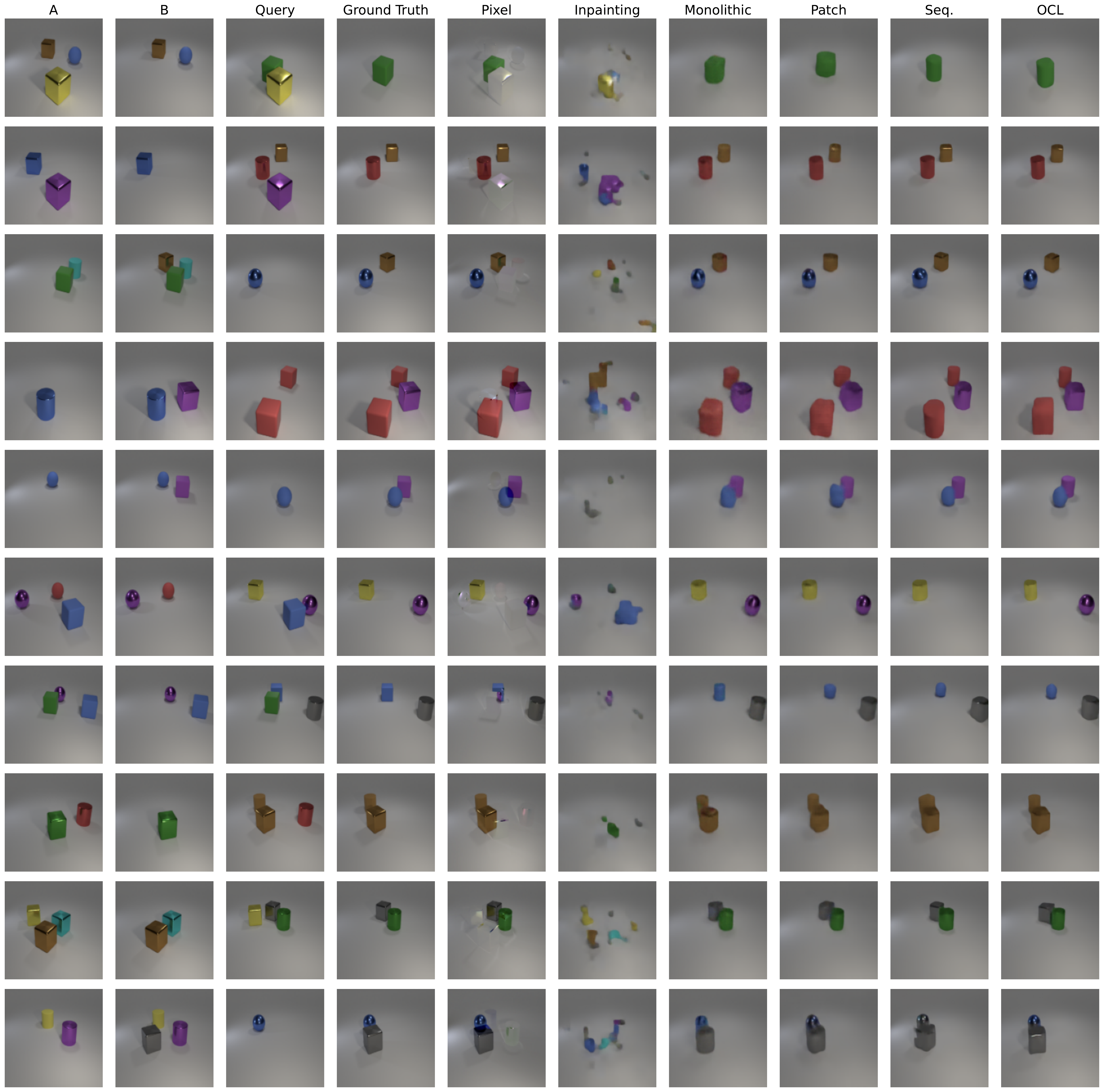}
    \caption{Primitive task extrapolation for the CLEVr Objects benchmark.}
    \label{fig:clevr_source_target}
\end{figure}

Figs.~\ref{fig:shapes3d_source_target}-\ref{fig:clevr_source_target} show generated examples in the primitive task 
extrapolation regime by different agents.

\subsection{Composite-Task Extrapolation:}

\label{app:composite_task}

\begin{figure}[!hbt]
    \centering
    \includegraphics[width=\linewidth]{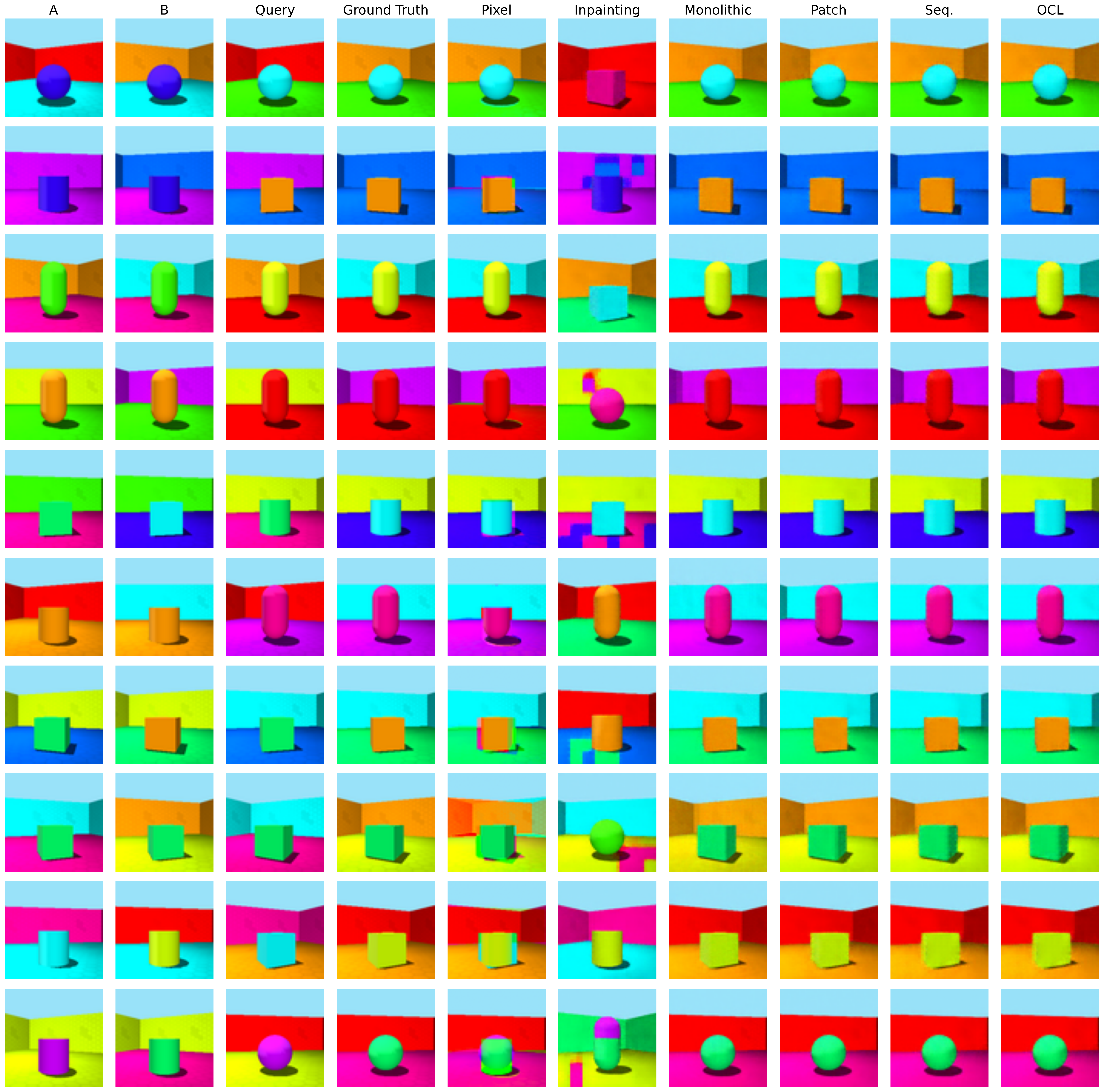}
    \caption{Two-composite task results on the 3D Shapes benchmark.}
    \label{fig:shapes3d_composite_2}
\end{figure}

\begin{figure}[!hbt]
    \centering
    \includegraphics[width=\linewidth]{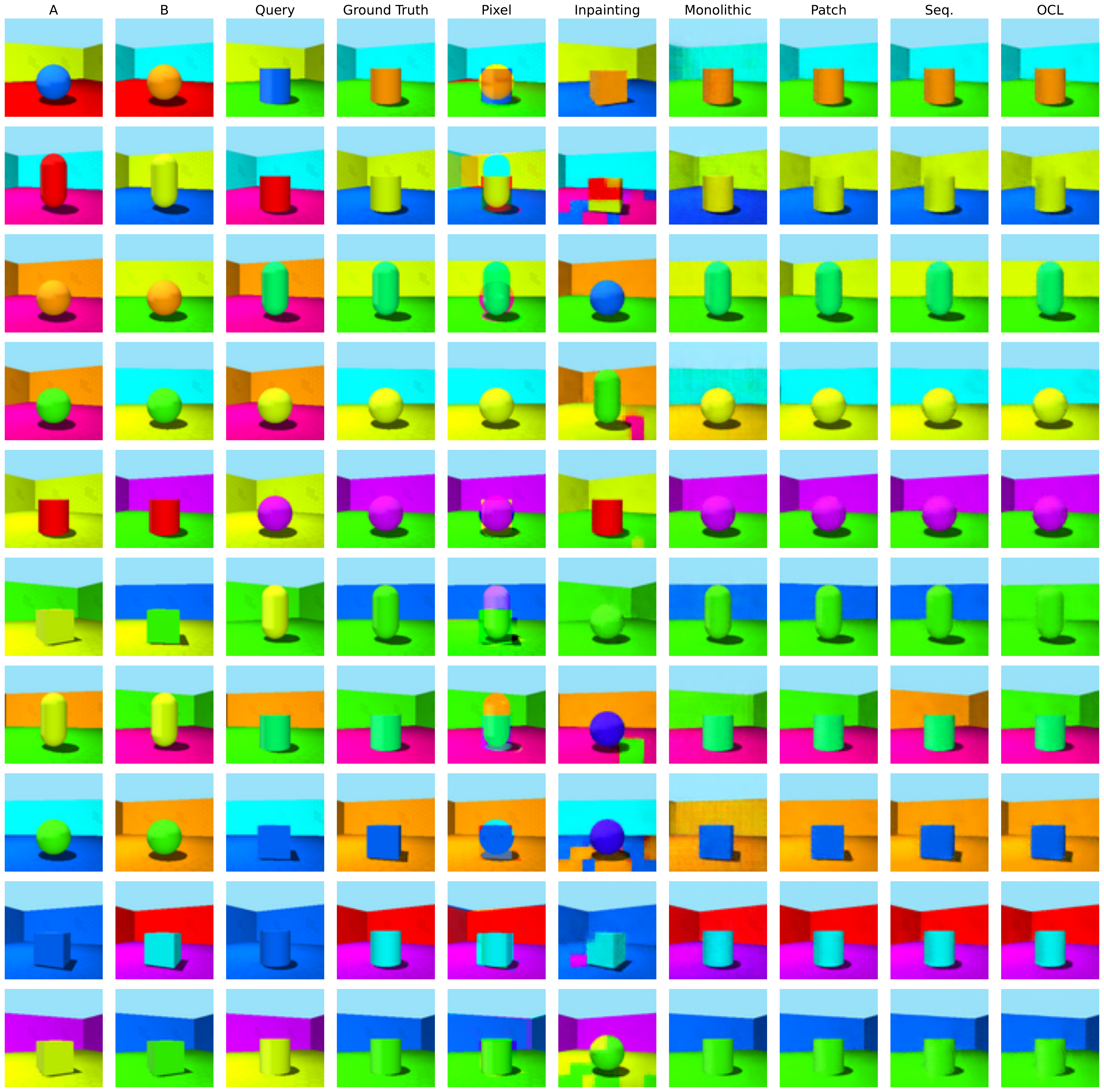}
    \caption{Three-composite task results on the 3D Shapes benchmark.}
    \label{fig:shapes3d_composite_3}
\end{figure}

\begin{figure}[!hbt]
    \centering
    \includegraphics[width=\linewidth]{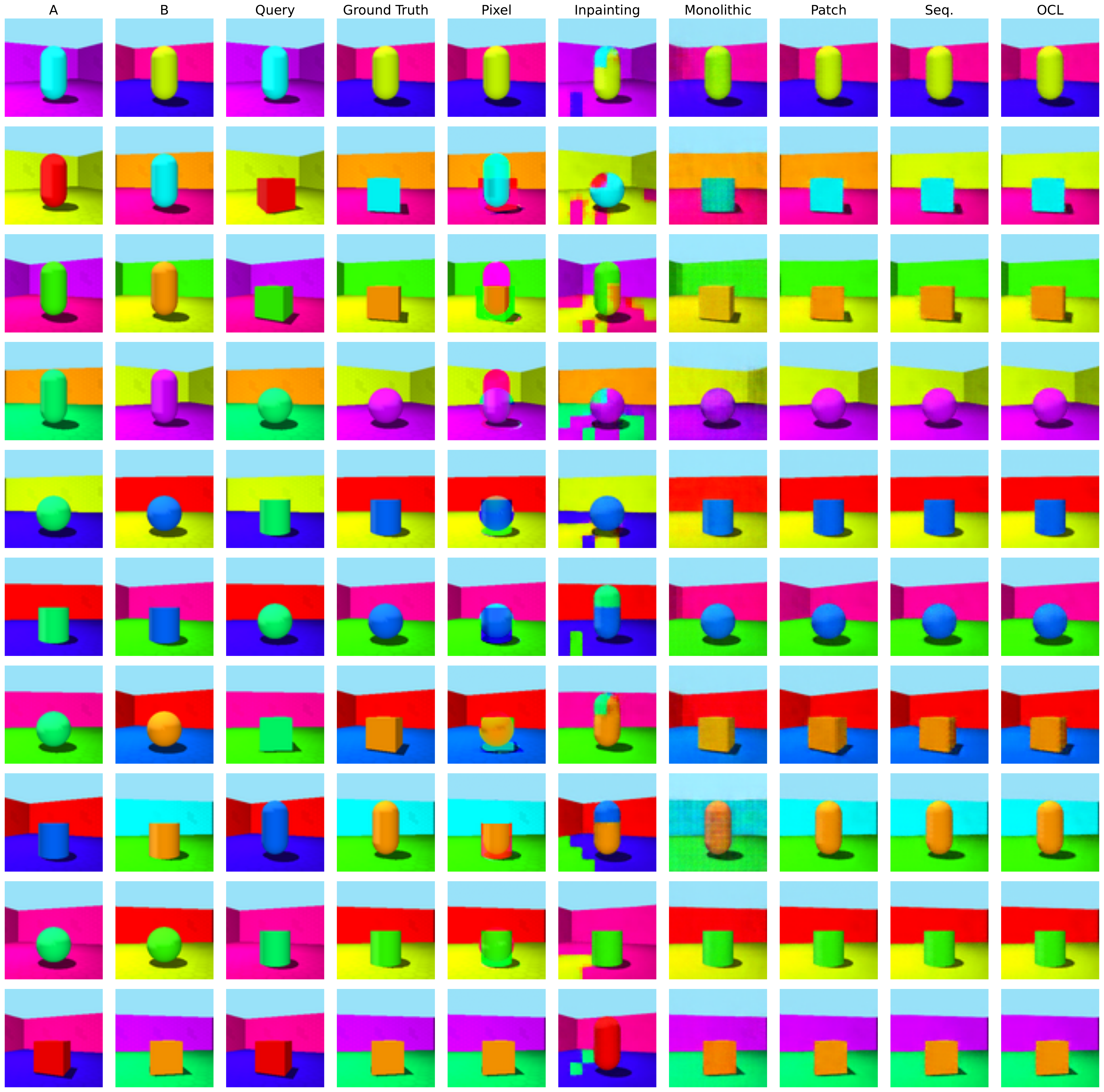}
    \caption{Four-composite task results on the 3D Shapes benchmark.}
    \label{fig:shapes3d_composite_4}
\end{figure}

\begin{figure}[!hbt]
    \centering
    \includegraphics[width=\linewidth]{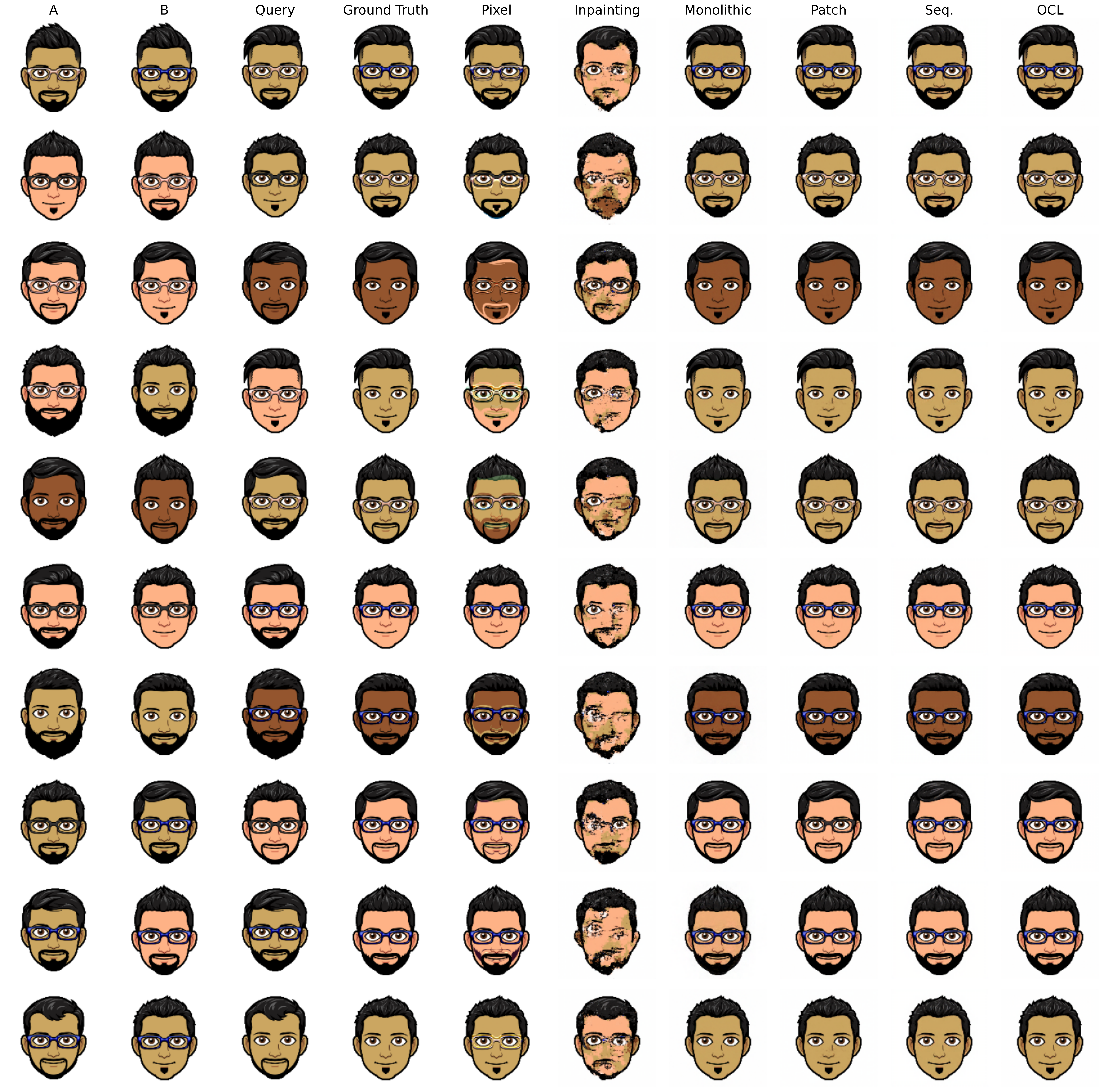}
    \caption{Two-composite task results on the BitMoji Faces benchmark.}
    \label{fig:bitmoji_composite_2}
\end{figure}

\begin{figure}[!hbt]
    \centering
    \includegraphics[width=\linewidth]{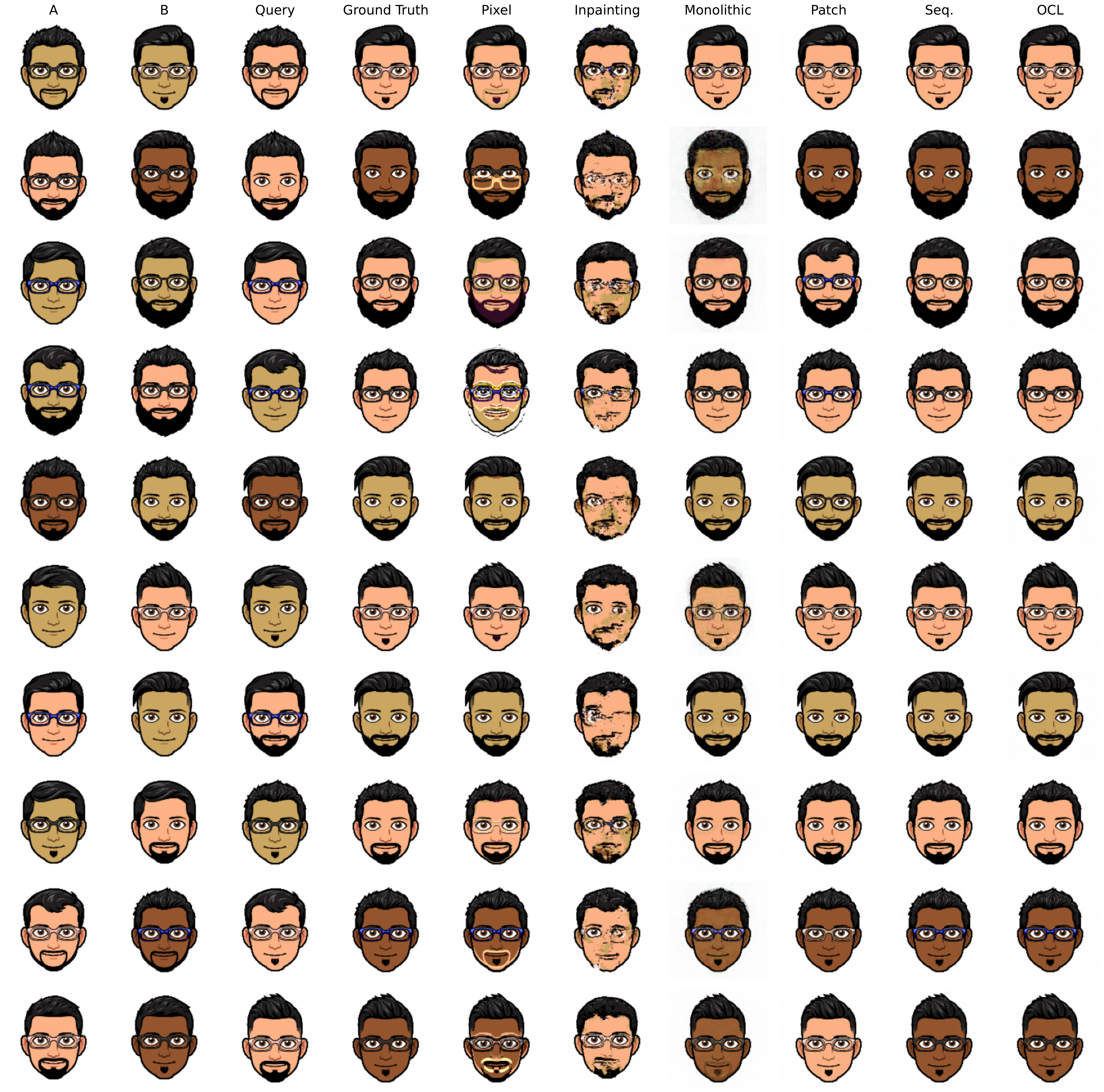}
    \caption{Three-composite task results on the BitMoji Faces benchmark.}
    \label{fig:bitmoji_composite_3}
\end{figure}

\begin{figure}[!hbt]
    \centering
    \includegraphics[width=\linewidth]{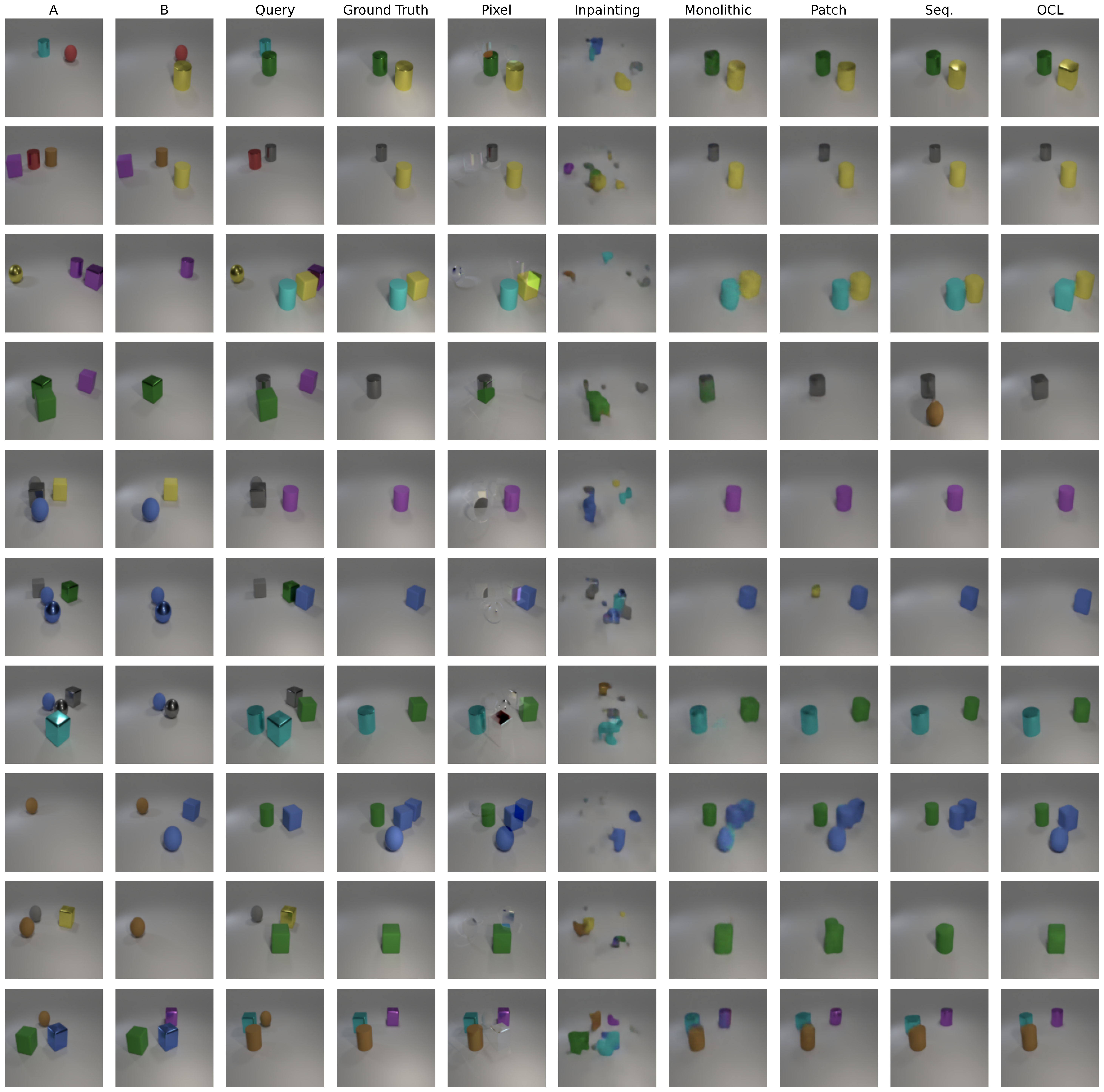}
    \caption{Two-composite task results on the CLEVr Objects benchmark.}
    \label{fig:clevr_composite_2}
\end{figure}

\begin{figure}[!hbt]
    \centering
    \includegraphics[width=\linewidth]{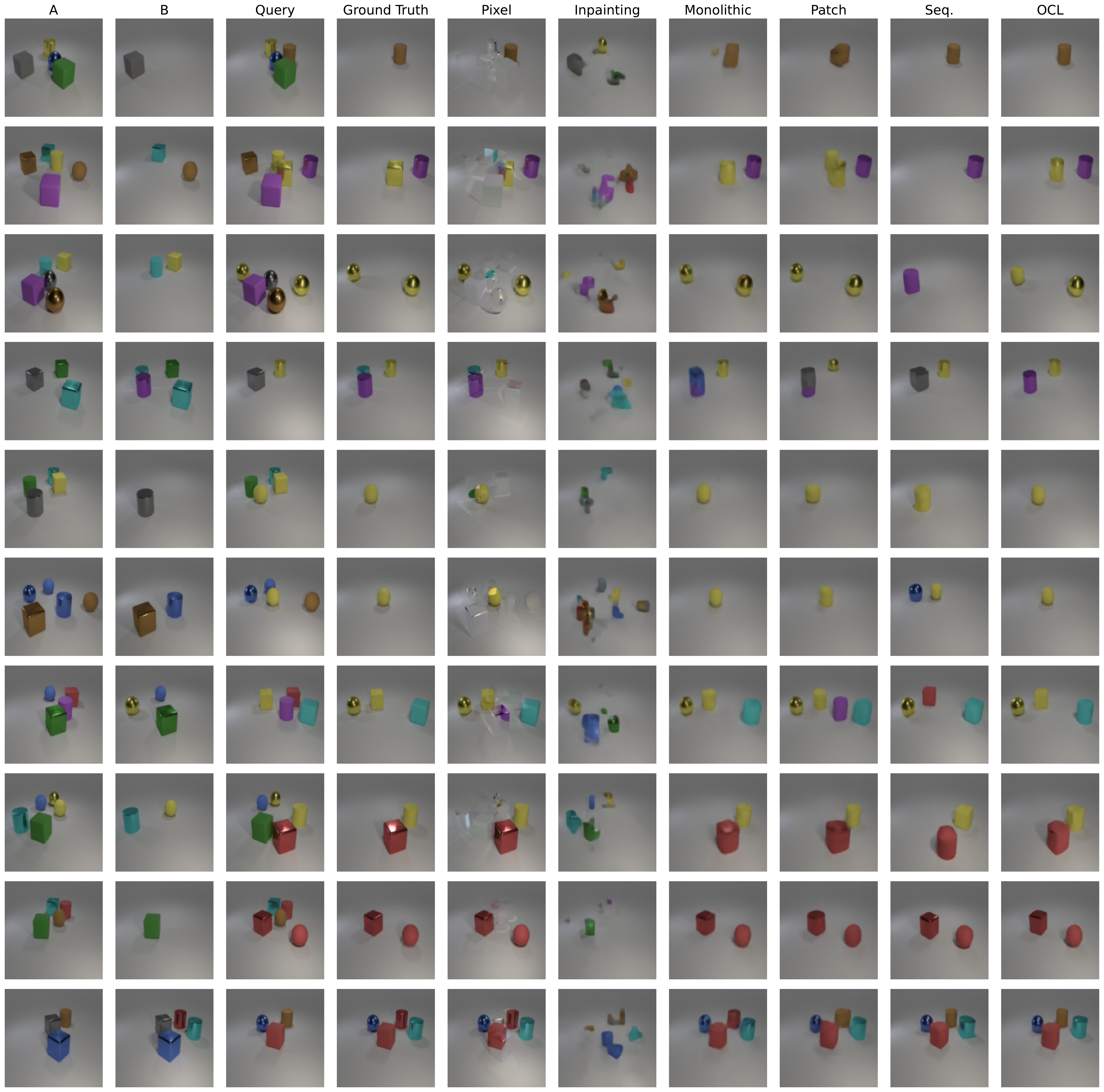}
    \caption{Three-composite task results on the CLEVr Objects benchmark.}
    \label{fig:clevr_composite_3}
\end{figure}

Figs.~\ref{fig:shapes3d_composite_2}-\ref{fig:clevr_composite_3} show generated examples in the composite tasks 
extrapolation regime by different agents.

\section*{Broader Impact}

This work demonstrates the potential for analogy-making to enable an in-context understanding of composition rules 
over visual stimuli. The benchmarks and meta-learning framework presented in this article provide a foundation for 
further exploration, which could lead to the development of more efficient and effective models for image generation 
and other visual tasks. While the current benchmark has no sampling bias over primitives, scaling up the models and the dataset can lead to 
biased outputs and the generation of counterfactual images capable of fooling humans.  

\end{document}